\pdfoutput=1

\documentclass[11pt]{article}

\usepackage{acl}

\usepackage{times}
\usepackage{latexsym}
\usepackage{adjustbox}
\usepackage[T1]{fontenc}
\usepackage[T1,T2A]{fontenc}

\usepackage[utf8]{inputenc}

\usepackage{microtype}

\usepackage{lipsum} 
\usepackage{arabtex}

\usepackage[english, vietnamese, russian]{babel}
\usepackage{utf8}
\setcode{utf8}
\usepackage{inconsolata}
\usepackage{microtype}
\usepackage{graphicx}
\usepackage{import}
\usepackage{layout}
\usepackage{tabularx, makecell}
\usepackage{booktabs}
\usepackage{mathrsfs}
\usepackage{amssymb} 
\usepackage{url}
\usepackage{hyperref}
\usepackage{graphicx}
\usepackage{xspace,paralist}
\usepackage{times,latexsym}
\usepackage{amsmath}
\usepackage{appendix}
\usepackage{comment} 
\usepackage{enumitem}
\usepackage{makecell}
\usepackage{multirow}
\usepackage{xcolor}
\usepackage{arydshln}
\usepackage{cleveref}
\usepackage{todonotes}
\usepackage{longtable,supertabular}
\usepackage[russian,english]{babel}
\usepackage[T2A,T1]{fontenc}

\usepackage{amssymb}
\usepackage{pifont}
\usepackage{CJKutf8}
\newcommand{\cn}[1]{\begin{CJK*}{UTF8}{gbsn}#1\end{CJK*}}
\newcommand{\cmark}{\ding{51}}%
\newcommand{\xmark}{\ding{55}}%

\newcommand{\tabref}[2][]{Table#1~\ref{#2}\xspace}
\newcommand{\figref}[1]{Figure~\ref{#1}\xspace}

\newcommand{\appref}[1]{Appendix~\ref{#1}\xspace}

\newcommand{\model}[1]{\text{#1}\xspace}

\newcommand{\claude}{\model{Claude}}
\newcommand{\chatglmfour}{\model{ChatGLM-4-9B-chat}}

\newcommand{\gptfouro}{\model{GPT-4o}}
\newcommand{\qwenturbo}{\model{Qwen-turbo}}
\newcommand{\qwentwo}{\model{Qwen2}}

\newcommand{\ecot}{\model{ECoT}}

\usepackage{CJKutf8}
\usepackage[utf8]{inputenc}

\NewDocumentCommand{\revanth}
{ mO{} }{\textcolor{blue}{\textsuperscript{\textit{Revanth}}\textsf{\textbf{\small[#1]}}}}

\title{Is Human-Like Text Liked by Humans? \\Multilingual Human Detection and Preference Against AI}

\author{Yuxia Wang\textsuperscript{1,7},
Rui Xing\textsuperscript{1}, 
Jonibek Mansurov\textsuperscript{1},
Giovanni Puccetti\textsuperscript{5}, 
\\\bf
Zhuohan Xie\textsuperscript{1}, 
Minh Ngoc Ta\textsuperscript{9}, 
Jiahui Geng\textsuperscript{1}, 
Jinyan Su\textsuperscript{1, 10}, 
Mervat Abassy\textsuperscript{14},
\\\bf
SaadElDine Eletter\textsuperscript{1, 14}, 
Kareem Elozeiri\textsuperscript{11}, 
Nurkhan Laiyk\textsuperscript{1}, 
Maiya Goloburda\textsuperscript{1}, 
\\\bf
Tarek Mahmoud\textsuperscript{1},
Raj Vardhan Tomar\textsuperscript{13},
Alexander Aziz\textsuperscript{15}, 
Ryuto Koike\textsuperscript{16},
\\\bf
Masahiro Kaneko\textsuperscript{1,16}, 
Artem Shelmanov\textsuperscript{1},
Ekaterina Artemova\textsuperscript{6}, 
Vladislav Mikhailov\textsuperscript{4}, 
\\\bf
Akim Tsvigun\textsuperscript{2,3}, 
Alham Fikri Aji\textsuperscript{1}, 
Nizar Habash\textsuperscript{1,8}, 
Iryna Gurevych\textsuperscript{1,12}, 
Preslav Nakov\textsuperscript{1}
\\
\textsuperscript{1}MBZUAI \quad
\textsuperscript{2}Nebius AI \quad 
\textsuperscript{3}KU Leuven \quad
\textsuperscript{4}University of Oslo 
\textsuperscript{5}ISTI-CNR\quad
\textsuperscript{6}Toloka AI
\\
\textsuperscript{7}INSAIT, Sofia University “St. Kliment Ohridski” \quad
\textsuperscript{8}New York University Abu Dhabi
\\
\textsuperscript{9}BKAI Research Center, Hanoi University of Science and Technology \quad
\textsuperscript{10}Cornell University
\\
\textsuperscript{11}Zewail City of Science and Technology \quad
\textsuperscript{12}TU Darmstadt \quad
\textsuperscript{13}CIC, University of Delhi  
\\
\textsuperscript{14}Alexandria University \quad
\textsuperscript{15}University of Florida \quad
\textsuperscript{16}Institute of Science Tokyo
\\
\href{yuxia.wang@mbzuai.ac.ae}{\{yuxia.wang, preslav.nakov\}@mbzuai.ac.ae}
}

\begin{document}
\selectlanguage{english}
\maketitle

\begin{abstract}
Prior studies have shown that distinguishing text generated by Large Language Models (LLMs) from human-written one is highly challenging for humans, and often no better than random guessing. To verify the generalizability of this finding across languages and domains, we perform an extensive case study to identify the upper bound of human detection accuracy. Across 16 datasets covering 9 languages and 9 domains, 19 annotators achieved an average detection accuracy of 87.6\%, thus challenging previous conclusions. We find that major gaps between human and machine text lie in concreteness, cultural nuances, and diversity. Prompting by explicitly explaining the distinctions in the prompts can partially bridge the gaps in over 50\% of the cases. However, we also find that humans do not always prefer human-written text, particularly when they cannot clearly identify its source. 
We release our dataset, the human labels, and the annotator metadata at \url{https://github.com/xnlp-lab/HumanEval-MGT}.

\end{list}
\end{abstract}

\section{Introduction}
Recent technological developments have advanced the generative artificial intelligence (AI) models, such as GPT-*, Gemini, Claude, and Llama~\cite{OpenAI2023GPT4TR, team2023gemini, Claude3, llama3}, thus aggressively blurring the lines between human and AI capabilities.
How often can state-of-the-art (SOTA) LLMs fool human evaluators, i.e.,~passing the Turing test? Can human evaluators correctly predict the origins of encountered content (human vs. machine), safeguarding against the potential misuse of LLMs?

Several studies have explored whether humans can distinguish between content generated by a human vs.\ a machine. Unsurprisingly, the findings depend in part on the capability of generative models. 
Early studies testing less advanced AI models found that human evaluators could indeed detect the difference~\cite{van-der-lee-etal-2019-best}. 
However, in studies using LLMs like GPT-3.5-turbo, human evaluators are frequently close to random chance~\cite{guo-etal-2023-hc3, jimpei2023creative, liam2023reakfake, chein2024human, wang2024m4gtbench}, particularly for sporadic users of LLMs. 

While covering different domains, e.g.,\ academic paper abstracts, Wikipedia paragraphs, question-answering responses~\cite{wang-etal-2024-m4, wang2024m4gtbench}, news and review comments~\cite{chein2024human}, most studies focused on GPT-3.5-turbo and English texts (see \tabref{tab:human-mgt-detection-survey}). Generally, fewer than 300 examples were evaluated by experts (LLM researchers or frequent users) in such studies. 
\begin{table*}[t!]
\tabcolsep3pt
  \centering
  \footnotesize
  \resizebox{\textwidth}{!}{
  \begin{tabular}{lcc llr ccc r}
    \toprule
    \textbf{} & \textbf{} & \textbf{} & \multicolumn{3}{c}{\textbf{MGT}} & \multicolumn{3}{c}{\textbf{Human Annotators}} & \textbf{Avg.Acc/}  \\
    \textbf{Study/Paper} & \textbf{Task} & \textbf{Lang} & \textbf{Models} & \textbf{Domain} & \textbf{\#Sample} & \textbf{Where and Size} & \textbf{Native} & \textbf{Layman} & \textbf{Range} \\
    \midrule 
    \multicolumn{10}{c}{\textbf{\textit{Before OpenAI released GPT-3.5-turbo in November 2022}} } \\\midrule
    \citet{garbacea-etal-2019-judge} & 2-class & English & 
     \begin{tabular}{@{}l@{}} 12 sequence \\ models pre 2019 \end{tabular}   & online product reviews & 3,600 & AMT-900 & -- & \cmark & 66.61\% \\\midrule
    \citet{ippolito-etal-2020-automatic} & 2-class & English & GPT2-large & web text & 150 & AMT, uni students & \cmark & \xmark & 71.4\% \\
    \citet{clark-etal-2021-thats} & 2-class & English & GPT2, GPT3 & stories, news articles, recipes & 3,900 & AMT-780 & \cmark & \cmark & 50-58\% \\  \midrule
    \begin{tabular}{@{}l@{}} \citet{shamardina2022findings}\\ \citet{shamardina2025coat} \end{tabular}& 2-class & Russian & \begin{tabular}{@{}l@{}} 14 decoder-only,\\ encoder-decoder\\ models \end{tabular} & \begin{tabular}{@{}l@{}} Wikipedia, news, social\\ media, RNC,diaries,\\ strategic documents \end{tabular} & 2,500 & Toloka & \cmark & \cmark & 66.66\% \\
    \midrule
    \multicolumn{10}{c}{\textbf{\textit{After OpenAI GPT-3.5-turbo}} } \\\midrule
    \citet{guo-etal-2023-hc3} & 2-class & English & GPT-3.5 & Wikipedia, QA & 150 & 8 experts + 9 layman & \xmark & \cmark & 48-90\%\\\midrule
    \citet{guo-etal-2023-hc3} & 2-class & Chinese & GPT-3.5 & Wikipedia, QA & 210 & 8 experts + 9 layman & \cmark & \cmark & 54-93\%\\\midrule
    \citet{chein2024human} & 2-class & English & GPT-3.5-turbo & news & 96 & 203 Prolific & \cmark & \cmark & 57\%\\\midrule
    \citet{chein2024human} & 2-class & English & GPT-3.5-turbo & social media comments & 96 & 203 Prolific & \cmark & \cmark & 78\%\\\midrule
    \citet{wang-etal-2024-m4} & 2-class & English & GPT-3.5-turbo & Reddit and arXiv abstract & 100 & 6 experts & \xmark & \xmark & 41-94\%\\\midrule
     \citet{liam2023reakfake} &  \begin{tabular}{@{}l@{}} boundary \\ detection \end{tabular}     & English & 
     \begin{tabular}{@{}l@{}} GPT2 (-XL)\\ CTRL \end{tabular} 
      & 
      \begin{tabular}{@{}l@{}} news, stories,\\ recipes, speeches \end{tabular}
       & 7,257 & RoFT game player & -- & \cmark & 23.4\% \\\midrule
    \citet{wang2024m4gtbench} & 4-class & English & 
    \begin{tabular}{@{}l@{}} davinci-text-003\\ GPT-3.5-turbo\\ Cohere\\ Dolly-v2\\ BLOOMz\\ GPT-4  \end{tabular}  
    & 
    \begin{tabular}{@{}l@{}} Wikipedia, \\ WikiHow, \\Reddit,\\ arXiv abstract,\\ peer review \end{tabular}
     & 140 & 4 experts & \xmark & \xmark & 10.4-27.4\% \\
    \bottomrule
  \end{tabular}
 }
  \caption{\emph{Task} refers to MGT detection task formulation. \emph{Experts} refers to individuals who are frequent users of LLMs, \emph{laymen} are people who have never heard of or used LLMs or have rarely used them, \emph{AMT} is Amazon Mechanical Turk. \emph{RNC} is the Russian national corpus.} 
  \label{tab:human-mgt-detection-survey}
\end{table*}
With the advancement of newer LLMs such as GPT-4o, Claude-3.5-Sonnet, and Llama4, we ask whether these findings generalize to them, to other languages, and to native expert annotators? What are the \textbf{upper bounds of human detection} performance across languages, domains and new LLMs? 

To answer these questions, we perform a comprehensive case study over 16 datasets spanning 9 languages, 9 domains, and 11 SOTA LLMs.
We focus on the following four research questions: 

(\emph{i})~\emph{How well can human annotators distinguish human-written text from SOTA LLMs generations?} 

(\emph{ii})~\emph{What are the notably detectable linguistic qualities in human- and AI-authored texts that shape decisions about text origin?} 

(\emph{iii})~\emph{Can a prompting strategy fill the gap between human and machine?} and 

(\emph{iv})~\emph{Do humans always prefer human-authored text?}

Our comprehensive analysis exposes the major gaps between current LLMs and human expectations. Our study challenge previous findings in terms of random-guess detection capability for humans and suggesting directions for future LLM development. 
%
Our contributions are as follows:
\begin{itemize}
    \item Our extensive case study shows that expert annotators achieve detection accuracy of \textbf{87.6\%} on average over $\sim$9K examples, demonstrating that humans can identify LLM outputs and safeguard against LLM misuse. We identified a spectrum of language- and domain-specific distinguishable features that distinguish human-written from machine-generated text, and we summarized five generalized distinctions across languages, which clearly expose the limitations of current LLM outputs and indicate directions for improvements. 
    \item Prompting by explicitly explaining the gap patterns between human text and LLM generations in instructions can either fully or partially address issues for over 50\% of the cases. However, cultural nuances, diversity in length, structure, and sentiment remain challenging.
    \item Humans do not always like human-written text: preferences are evenly split between human vs. machine, with a strong tendency to favor machine-generated text (MGT) when annotators can not clearly identify the origin of a given text. 
    \item We release 17k original and 32k improved MGTs, labels of 13.5k human detection instances, 5k preferences and 1.6k fill-in-the-gap survey, and metadata for 19 annotators to facilitate future work on human vs. MGT detection, preference, and their relationship with individual characteristics.  
\end{itemize}

\section{Case Study}
\label{sec:data-annotation}

\begin{table*}[t!]
\centering
\tabcolsep3pt
\resizebox{\textwidth}{!}{\small
\begin{tabular}{l p{3.4cm} p{7.3cm} p{2.2cm} p{3.6cm}}
\toprule
\textbf{Setting ID} & \textbf{Input} & \textbf{Task Description} & \textbf{Outputs} & \textbf{Applicable Scenarios} \\
    \midrule 
    \multirow{2}{3cm}{\textbf{I. Single-Binary}} & \texttt{hwt} or \texttt{mgt} & Given a piece of text, identify whether it is written by human? & \textbf{A.}~Yes, human;\phantom{x} \textbf{B.} No, machine & parallel data is not necessary. \\
    \midrule
    \multirow{3}{3cm}{\textbf{II. Pair-Binary}} & (\texttt{hwt}, \texttt{mgt}) \textbf{or} (\texttt{mgt}, \texttt{hwt}) & Given a pair of (text1, text2), identify which one is human-written? Either text1 or text2 must be \texttt{hwt}, and another is \texttt{mgt} randomly sampled from $mgt_i$. & \textbf{A.}~text1;\phantom{xxxxxxx}  \textbf{B.}~text2 & parallel data is available. \\
    \midrule
    \multirow{3}{3.5cm}{\textbf{III. Triplet-Three-Class}} & (\texttt{hwt}, \texttt{mgt}$_1$, \texttt{mgt}$_2$)  & Given a set of texts (text1, text2, text3), identify which one is human-written? One of the text1, text2 and text3 must be \texttt{hwt}, and others are \texttt{mgt} randomly sampled from $mgt_i$. & \textbf{A.}~text1;\phantom{xxxxxxx}  \textbf{B.}~text2;\phantom{xxxxxxx}  \textbf{C.}~text3 & parallel human and multiple LLM generations are collected to make comparisons. \\
    \midrule
    \multirow{2}{3cm}{\textbf{IV. Pair-Four-Class}} & (\texttt{hwt}, \texttt{mgt}) \textbf{or} (\texttt{mgt}, \texttt{hwt}) \textbf{or} (\texttt{hwt}, \texttt{hwt}) \textbf{or} (\texttt{mgt}, \texttt{mgt}) & Given a pair of (text1, text2), identify which one is human-written? Both text1 and text2 can be \texttt{hwt}, and can be \texttt{mgt}. & \textbf{A.}~text1; \textbf{B.}~text2; \textbf{C.}~none; \textbf{D.} both & parallel data is not necessary. \\
    \bottomrule
\end{tabular}
}
\caption{The four human detection settings: the setting name refers to input-output options, pair/binary means the input is a pair of texts and the goal is to predict binary labels, and whether the text is human-written (Yes or No).}
\label{tab:detection-settings}
\end{table*}
\subsection{Human Detection Setups}
\label{sec:anno-setup}

Previous studies presented the difficulty humans face in distinguishing SOTA LLM-generated content from human-written text, often resulting in a random guess (see more in \appref{sec:relatedwork}). However, most evaluations focused on English and generations by GPT-3.5-turbo, leaving the detectability of MGT in other languages and LLMs uncertain.

To verify whether this observation can generalize to other languages and more advanced LLMs, we collected LLM generations based on 16 datasets across 9 languages and 9 domains to evaluate a broad spectrum of writing styles. 

A group of 19 native speakers who are either LLM researchers or practitioners performed human evaluations, investigating (\emph{i})~whether expert humans can correctly discern human vs.\ AI outputs, and (\emph{ii})~whether humans prefer fellow human answers or LLM responses.

\subsection{Task and Dataset}

\paragraph{MGT Detection} The goal is to identify whether the text was written by a human or generated by models given a single text, or to recognize which text is written by a human given a pair of texts: one human-written and one machine-generated.

We collect datasets from common domains (community QA, news, tweets, government reports), and domains requiring high-integrity LLM applications, including educational and academic contexts, such as accurate knowledge verification in Wikipedia-style texts, and identifying the authorship of student essays and peer reviews. 

For each language and dataset, we sampled 300-600 human texts and then generated corresponding machine text using two SOTA LLMs: a multilingual model (e.g.,\ GPT or Claude series) and a language-specific model (e.g.,\ ChatGLM or Qwen for Chinese and AceGPT for Arabic), to analyze the impact of different LLMs on detection performance, particularly for non-English languages.
\appref{sec:datasets} summarizes the 16 datasets, including statistics, prompts, and collection details.

\paragraph{Annotation Settings}
To mimic real-world MGT detection scenarios, we set up four human evaluation settings. 
Given human-written text and machine-generated text, representing by \texttt{hwt} and \texttt{mgt} respectively, human annotators are asked to identify which text was written by a human and which by a LLM. Note that \texttt{mgt} can be generated by multiple different LLMs, referring to as \texttt{mgt}$_\texttt{i}$, where \texttt{i}$\in \mathbb{N}$. 

As shown in \tabref{tab:detection-settings}, according to the input and the output options, we categorize detection settings as I. \emph{single-binary}, II.  \emph{pair-binary}, III. \emph{triplet-three-class}, and IV. \emph{pair-four-class}. 
For a single text input, either \texttt{hwt} or \texttt{mgt}, the goal is to recognize whether the text was written by a human, by answering just Yes or No. This is suitable for the scenario where for the human text there is no necessarily a corresponding machine-generated text, and thus they can be collected from different sources and for different topics, such as Arabic tweets.
Given a pair of texts (text1, text2), a binary output is much easier than the four-class detection setting. The \emph{pair-binary} setting asks that either text1 or text2 is \texttt{hwt}, and the other one is \texttt{mgt}, while the \emph{pair-four-class} setting has no restrictions: each of text1 and text2 can be \texttt{hwt} or \texttt{mgt}, regardless of the label of the other text. 
Sometimes, we want to compare human text to generations from different LLMs, in which case, we apply III, which we limit to a three-class detection: human vs.\ LLM$_1$ vs.\ LLM$_2$.

Overall, II and III are applicable in scenarios where a human text has a corresponding machine-generated counterpart. I and IV can be utilized when the human- and machine-generated texts are non-corresponding.
Settings I and III are commonly used in other studies, as well as this work.
If annotators are exposed to some human- and machine-written texts prior to the detection task, we refer to it as a few-shot setting; otherwise, it is considered zero-shot.

\begin{table*}[t!]
    \centering
    \small
    \resizebox{\textwidth}{!}{
    \begin{tabular}{llr lll c}
    \toprule
    \textbf{Language} & \textbf{Source/Model} & \textbf{\#Example} & \textbf{\#Annotator} & \textbf{Anno Setup} & \textbf{Shot} & \textbf{Avg. Acc} \\
    \midrule
    \multirow{4}{*}{Arabic} 
    & Dialect Tweet & 900 & 1 BSc. & I. Single-binary & Zero & 50.1  \\
    & EASC Summary & 100 & 1 BSc. & II. Pair-binary & Zero & 82.0  \\
    & Youm7 News & 1,000 & 1 BSc. & II. Pair-binary & Zero & 92.7  \\
    & SANAD News & 100 & 1 BSc. & II. Pair-binary & Zero & 100.0  \\
    \midrule
    \multirow{6}{*}{Chinese} 
    & Zhihu-QA (\gptfouro) & 428 & 1 M, 1 PhD, 3 PD & II. Pair-binary & Zero & 99.6 \\
    & Zhihu-QA (\gptfouro) & 160 & 1 MSc. & II. Pair-binary & Few  & 100.0  \\
    & Zhihu-QA (\qwenturbo) & 588 & 1 PhD, 1 Postdoc & II. Pair-binary & Zero & 98.0 \\
    & Student essay & 102 & 1 Postdoc & II. Pair-binary & Few  & 98.0  \\
    & Student essay & 600 & 1 PhD, 2 Postdoc & IV. Pair-four-class & Zero & 97.0 \\
    & Government Report & 500 & 1 Postdoc & III. Triplet-three-class & Few & 97.2  \\
    \midrule
    English & Peersum & 400 & 1 BSc. & II. Pair-binary & Few & 99.8   \\
    \midrule
    Hindi & News & 600 & 1 BSc. & II. Pair-binary & Few & 85.2   \\
    \midrule
    \multirow{3}{*}{Italian} 
    & DICE News (Anita) & 300 & 1 Postdoc & II. Pair-binary & Few & 88.0  \\
    & DICE News (Llama3-405B) & 300 & 1 Postdoc & II. Pair-binary & Few & 99.7  \\
    & DICE News (GPT-4o) & 300 & 1 Postdoc & II. Pair-binary & Few & 100.0  \\
    \midrule
    Japanese & News & 300 & 1 PhD, 1 Postdoc & II. Pair-binary & Zero & 86.4  \\
    \midrule
    Kazakh & Wikipedia & 300 & 2 MSc. & II. Pair-binary & Zero & 79.7   \\
    \midrule
    \multirow{2}{*}{Russian} 
    & News & 300 & 1 PhD & II. Pair-binary & Few & 100.0   \\
    & Academic summary & 300 & 1 PhD & I. Single-binary & Few & 80.0   \\
    \midrule
    \multirow{2}{*}{Vietnamese} 
    & Wikipedia & 600 & 1 BSc.  & II. Pair-binary & Zero & 50.7  \\
    & News & 600 & 1 BSc. & II. Pair-binary & Zero & 80.3  \\
    \midrule
    \bf Total & -- & 8,778 & 30 & & & \textbf{87.6}\\
    \bottomrule
    \end{tabular}
   }
    \caption{Human annotator detection accuracy over 16 datasets and 9 languages: we have a total of 30 annotation settings and 19 unique human annotators. The simple average accuracy of the human expert guesses is 87.6\%.}
    \label{tab:detection-acc}
\vspace{-5pt}
\end{table*}

\paragraph{Annotator Background}
We highlight that this study aims to explore the upper bound of human detection capability, rather than represent the full population distribution. Therefore, we did not recruit lay annotators who might have limited experience with LLMs and varying familiarity with machine-generated texts.

Instead, we conducted in-house annotation with six fourth-year BSc.\ with 1-2 years of research experience in MGT detection, three MSc., and six PhD. students, as well as four postdocs, all with NLP expertise and prior experience with LLM-generated texts. Annotators were native speakers of the respective languages and completed annotations independently (see \appref{sec:whyexperts} for details).

While using expert annotators may limit the generalizability of results to lay users, our goal is to establish expert performance ceilings. If trained experts struggle to detect machine-generated text, it is reasonable to expect lower performance from the general public. Lay annotator studies are important for practical applications, but do not substitute for upper-bound evaluation. Future work will include lay annotators to capture a broader distribution of human detection capabilities.

\begin{table}[t!]
    \centering
    \small
    \resizebox{\columnwidth}{!}{
        \begin{tabular}{lcccc}
            \toprule
            \textbf{Dialect} & \textbf{Human} & \textbf{\gptfouro} & \textbf{\qwentwo} & \textbf{All MGT} \\
            \midrule
            EGY & 52.0 & 53.3 & 58.7 & 56.0 \\
            MOR & 54.0 & 53.3 & 48.0 & 50.7 \\
            LEV & 69.3 & 14.7 & 58.7 & 36.0 \\
            GULF & 81.3 & 26.7 & 30.7 & 28.7 \\
            \bottomrule
        \end{tabular}
    }
    \caption{Arabic dialect tweet human detection accuracy over human vs. \gptfouro vs. \qwentwo-7.5B. Machine-generated text is harder than human text to discern. \gptfouro is harder than \qwentwo-7.5B.}
    \label{tab:arabic-dialect_tweet-accuracy}
\end{table}

\section{Human Detection}
\label{sec:human-detection-acc}
We performed an extensive case study on 9k instances across 9 languages to verify how difficult it is for native speakers to detect AI outputs in everyday domains.
\tabref{tab:detection-acc} demonstrates that human detection accuracy can reach 87.6\% on average.
Contrary to previous findings, \textbf{mgt} detection is not particularly difficult for native human experts. 

\subsection{Language}
Human detection accuracy exceeds 87.6\% for Chinese, English, Arabic, Italian and Russian, while it falls below this level for Vietnamese, Kazakh, Hindi, and Japanese. This discrepancy is largely due to the challenge of Wikipedia text.

\subsection{Domain}
Wikipedia ubiquitous presence in training datasets leads to Wikipedia-like \textbf{mgt} resembling human-written Wikipedia.
\tabref{tab:arabic-dialect_tweet-accuracy} shows that Arabic tweets also present challenges for detection due to their short length, and four dialects further increase the difficulty. Similar patterns appear in summaries, e.g.,\ for Arabic and Russian summaries, the expert detection accuracy is about 80\%.

This conversely highlights the ability of language models to generate high-quality human-like text in the domains of Wikipedia, tweets, and summaries. In contrast, substantial differences between machine-generated and human-written text persist in news articles, QA, student essays, and peer reviews, making them much easier to recognize for human experts.


\subsection{Generator} 

The impact of generators is associated with languages and domains. 
While there are minimal differences for Chinese, where accuracy is 100\% vs.\ 98\% for \gptfouro vs.\ \qwenturbo, there are sizable differences for Italian and Arabic.
Based on Italian DICE News, the same annotator detected generations by Anita (an Italian fine-tuned Llama3-8B), Llama3-405B, and \gptfouro, achieving accuracy of 88\%, 99\%, and 100\%, respectively.
Similarly, for Arabic tweets, \gptfouro's outputs are more similar to human text and thus more difficult to detect compared to those by \qwentwo, as clearly shown in \tabref{tab:arabic-dialect_tweet-accuracy}.

\subsection{Annotation Setting}
We conducted the majority of our annotations under setting II. \emph{pair-binary}: given a pair (\texttt{hwt}, \texttt{mgt}), it asks to identify which of the two texts is human-written. 
We assumed that more complex settings would be more challenging. For instance, IV. \emph{pair-four-class} should be harder than II, as each of the texts could be human- or machine-generated, independently of the other. I. \emph{single-binary} should be the hardest as it provides annotators with the least additional information: a single text without references to compare.

Yet, for Chinese student essays, the performance for IV does not degrade compared to II.
Similarly, for government reports in III. \emph{triplet-three-class}, where the annotators have to select the human-written text among three candidates, there was no degradation compared to II.  

I. \emph{single-binary} proves to be more difficult than II. \emph{pair-binary} for both Arabic and Russian. While domain differences do exist, e.g.,\ tweets vs.\ summary vs.\ news in Arabic, and news vs.\ summary for Russian, the substantial performance gap (>20\%) can still be partially attributed to the annotation settings.
Overall, comparing the results for I. vs.\ II., it is easier to distinguish machine-generated content when given a comparison pair, rather than for single answer. 

Yet, introducing text triplets or increasing the number of machine-generated or human-written texts had minimal impact on detection performance. Moreover, using few shots before detection boosted the confidence of the annotators, resulting in higher accuracy compared to zero-shot.
For datasets with high accuracy, before seeing labeled samples, the annotators found the distinction to be obvious. After seeing a few examples, the annotator was extremely confident in distinguishing human vs.\ machine text based on indicative features of MGT.

\subsection{Expert Annotators}
For the same language and dataset, individual annotator ability influences detection accuracy but not significantly. 
For instance, in Chinese Zhihu-QA (GPT-4o vs.\ human), five annotators achieved accuracies of 99\%, 99\%, 100\%, 100\%, and 100\%. Similarly, for Zhihu-QA (Qwen-turbo vs.\ human), two annotators obtained 99\% and 97\%. In student essays (IV. pair-four-class), three annotators recorded accuracies of 96\%, 96\%, and 99\%.
This may also result from the bias that all annotators are native speakers and expert-level LLM practitioners or researchers. Differences between individuals are minor in their cognitive abilities, language proficiency and domain knowledge.


\subsection{Distinguishable Factors}
\label{sec:dis-factor}
We summarize five remarkable distinguishable signals between machine-generated vs.\ human-written text across the 16 datasets and the 9 languages. \appref{sec:distinctionfactor} provides detailed distinctions between \texttt{hwt} and \texttt{mgt} for each language and dataset, where valuable expert insights into the limitations and gaps of current LLM outputs compared to human-written text can inform future model improvements.

\begin{itemize}
    \item \textbf{Human text is more informative and concrete.}
    Human-written text contains concrete numbers, specific names of people or institutions, exact places or dates, URLs, and other references, while machine-generated text tends to provide generic information, with little detail to support its statements. 
    
    \item \textbf{Machine-generated text lacks regional, cultural, and religious nuances.}
    For languages such as Arabic, Japanese, Hindi, Kazakh, and Chinese, human texts reflect the cultural and the religious nuances of the language, which is not true for machine-generated text.

    \item \textbf{Human-written text varies substantially in terms of length, structure, style, and sentiment.}
    Human texts show diversity and inconsistency with large deviations in length, structure, style and emotions, while machine-generated texts tend to use a formulaic structure and neutral/positive emotion. This can be partially attributed to LLMs rigorously following instructions, and thus losing on flexibility.

    \item \textbf{Machine-generated text has formatting.}
    MGTs are generally well-segmented with bullet points for better readability, while human-written texts are typically just large block of plain text, which may be due to human text collection and conversion. Moreover, machine-generated texts often use Markdown style, e.g.,\ \texttt{**} and \texttt{\#\#\#}, while human-written texts have typos, grammatical errors, hashtags, and other social media elements.

    \item \textbf{Machine-generated text shows a mixture of other languages.} Non-English language responses often contain some English parts, and sometimes mix other languages, e.g.,\ Japanese responses contain Chinese or Korean, particularly for less capable models. This is very rare in human text. 
\end{itemize}


\section{Can Prompting Fill in the Gap?}

\begin{figure}[t!]
    \centering
    \includegraphics[scale=0.38]{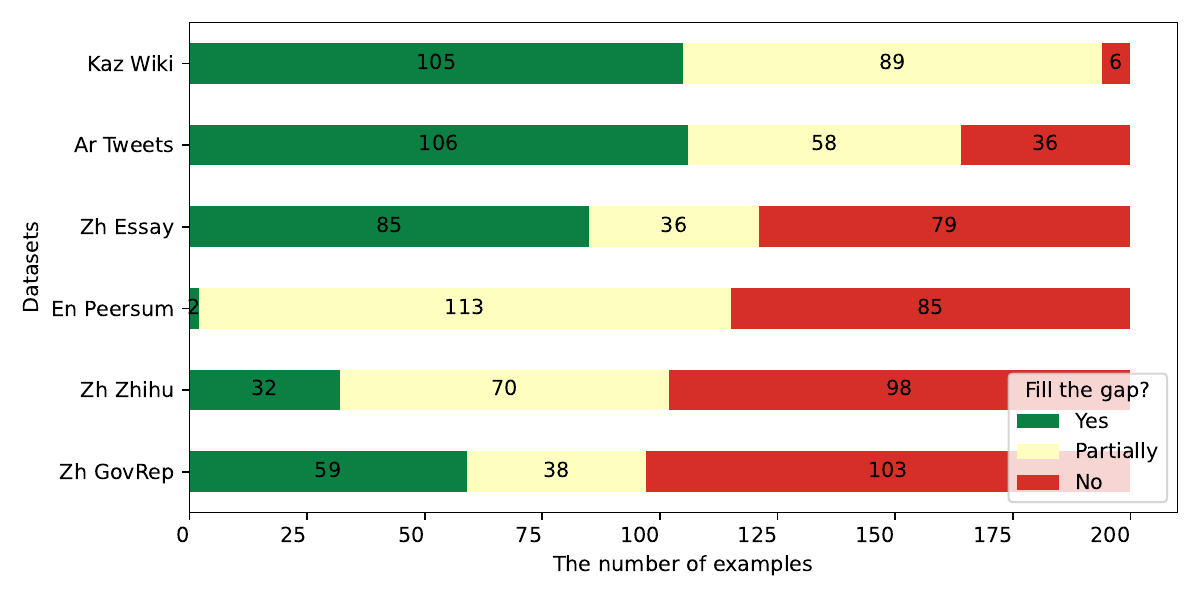} 
    \caption{Evaluating whether the new generations fill in the gap: Yes, Partially, or No.}
    \label{fig:dist-survey}
\end{figure}

Given that LLMs can closely follow instructions and their outputs are heavily influenced by the system and the user prompts, we investigated whether explicitly instructing them to mimic human style can help narrow the gap. 
Responding to the distinguishable factors summarized in \appref{sec:distinctionfactor} for each dataset, we asked the human annotators to craft new prompts, aiming to improve the generations and to reduce the gap between human-written and LLM-generated texts. 

This involved trying instructions that (\emph{i})~incorporate specific details and references, (\emph{ii})~avoid formulaic structures and formats, e.g.,\ bullet points and Markdown, and (\emph{iii})~generate texts of varying length, structure, and sentiment.
\appref{sec:fill-gap-prompts} presents both the original and the improved prompts for all datasets.

\textbf{Measurements:}
We re-generated the machine-generated parts of the text pairs, using the same models with improved prompts, and then sampled 200–600 examples from each dataset to assess whether and to what extent, the prompting strategy narrowed the gap between human-written and machine-generated texts. 

We used two approaches: (\emph{i})~intuition-oriented \textit{fill-the-gap survey}, where the original annotators evaluated whether the newly-generated text bridged the gap (Yes, No, or Partially), and (\emph{ii})~\textit{downstream detection}, where we compared the detection accuracy before and after applying the improved prompts.  A decline in detection accuracy indicated a reduced distinction between human and machine text, making the differentiation more difficult, and further revealing that prompting was effective. 
While the survey is inherently slightly subjective, it remains valuable, as human intuition plays a crucial role in evaluating the perceived quality of model outputs \citep{elangovan-etal-2024-considers}. By combining this intuition-driven assessment with quantitative detection results, we aim to draw more robust and convincing conclusions. 

\begin{figure}[t!]
    \centering
    \includegraphics[scale=0.4]{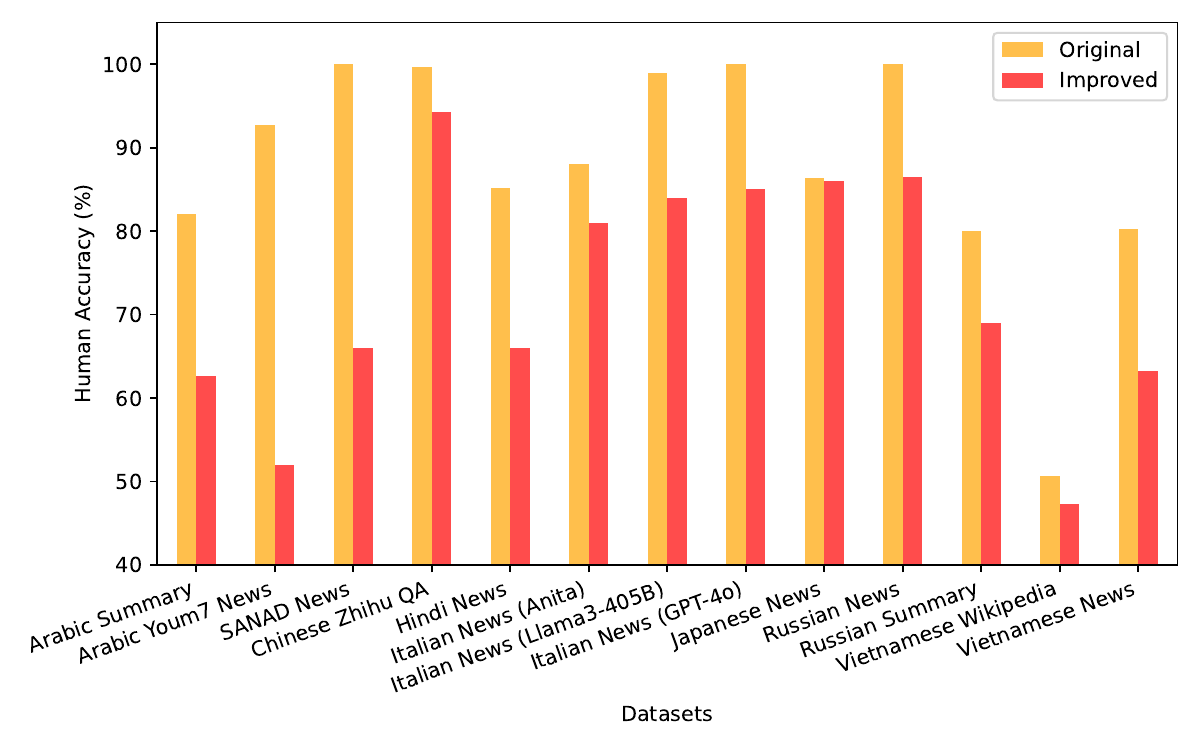}
    \caption{Human detection accuracy dropped from 87.6 to 72.5 for the original vs.\ the improved generations.}
    \label{fig:acc-diff}
\end{figure}




\begin{figure*}[t!]
    \centering
    \includegraphics[scale=0.55]{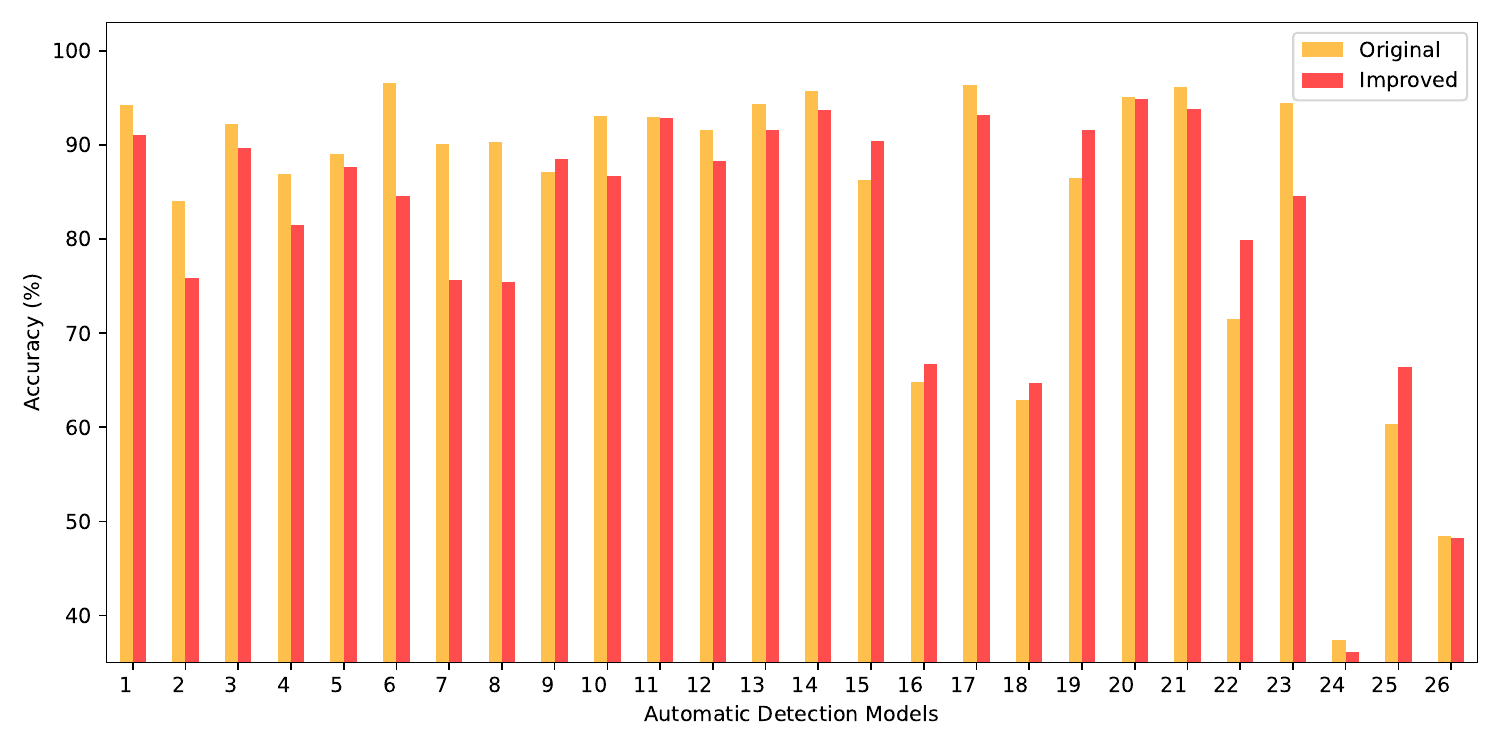}
    \caption{\textbf{Detection accuracy differences} of 26 automatic MGT detection approaches on original vs. improved.}
    \label{fig:auto-acc-diff}
\end{figure*}
\paragraph{Fill-the-gap Survey}
The original annotators who conducted the earlier detection were asked to evaluate whether the text generated by the new prompts addressed the gaps for each example, providing one of three labels: \emph{Yes}, \emph{No}, and \emph{Partially}.

The distributions across the six datasets in \figref{fig:dist-survey} shows that in about 50\% of the cases, the prompt adjustments were effective to  either fully or partially mitigate the gaps. Large improvements were observed for Kazakh Wikipedia and Arabic tweets.
For the former, the revised prompt reduced repetitive sentence patterns (more diverse), but the formulaic expressions were not entirely eliminated. New outputs also included more concrete information, such as dates and names, while the inclusion of culturally-nuanced details remained challenging. The newly-generated Arabic tweets could touch on relatable human topics and express genuine emotions tied to daily experiences; however, the frequently added irrelevant hashtags at the end of the tweets and a occasionally inappropriate optimistic tone made them easily identifiable as machine-generated. 

The annotators for English peer reviews noted that, despite the prompt adjustments, the models remained highly formulaic in their outputs, the length of the reviews remained relatively uniform, and the structure lacked variance. This may be due to the inherent nature of the peer review domain, while human reviews exhibited more variability in both length and structure, feeling more organic.
Similar issues remained for Chinese student essays (formulaic structure by using ``firstly, then, moreover, finally'' persisted) and government reports (certain repetitive phrases). See more details in \Cref{sec:chinese-distinguishable-signal}.
Overall, adjusting the prompts did fill some gaps, but cultural nuances, diversity of length, structure and phrases, and sentiment adaption to scenarios remained challenging.

See more in \appref{sec:fill-gap-prompts}.
With the intuitive survey results, we further conducted a second round of MGT detection on the new generations for objective results, involving both human evaluation and automated detection.

\paragraph{MGT Detection on Improved Text}
We performed human detection on 13 datasets under the same annotation setting as described in \tabref{tab:detection-acc}, with the exact same annotators.
We observed sizable accuracy declines across all datasets in \figref{fig:acc-diff}, with average accuracy dropping to 72.5\% (detailed results are in \appref{sec:fill-gap-prompts}).
This implies that the improved machine-generated text became more similar to the human-written text, making it harder to discern and thus resulting in lower detection accuracy.

We further analyzed the impact of the improved generations on automatic detection accuracies.
We collected a total of 17,017 texts using the original prompts and 32,487 texts using the improved prompts (detailed statistical distribution in \appref{app:automaticdetectiondata}).
We reproduced 26 MGT detection approaches presented in the COLING 2025 GenAI shared task~\cite{wang-etal-2025-genai} and evaluated them on the collected MGTs. As shown in \figref{fig:auto-acc-diff}, 19 methods exhibited lower accuracy on the newly generated texts, indicating that the texts produced using the improved prompts are more challenging to distinguish compared to those generated with the original prompts.
This suggests that prompting strategies can help bridge some gaps between machine-generated and human-written text when explicitly designed to mimic human writing style.

\section{Human-Like or Liked-by-Humans?}
We used the prompting strategy to bridge the gap between human and machine text, aiming to make machine outputs more human-like. However, do humans favor human-like text? 
Below, we examine human preferences among four options: (\emph{i})~human-written text, (\emph{ii})~machine-generated text using the original prompt, (\emph{iii})~machine-generated text using the improved prompt, or (\emph{vi})~none of the above. 

\begin{figure}[t!]
    \centering
    \includegraphics[scale=0.37]{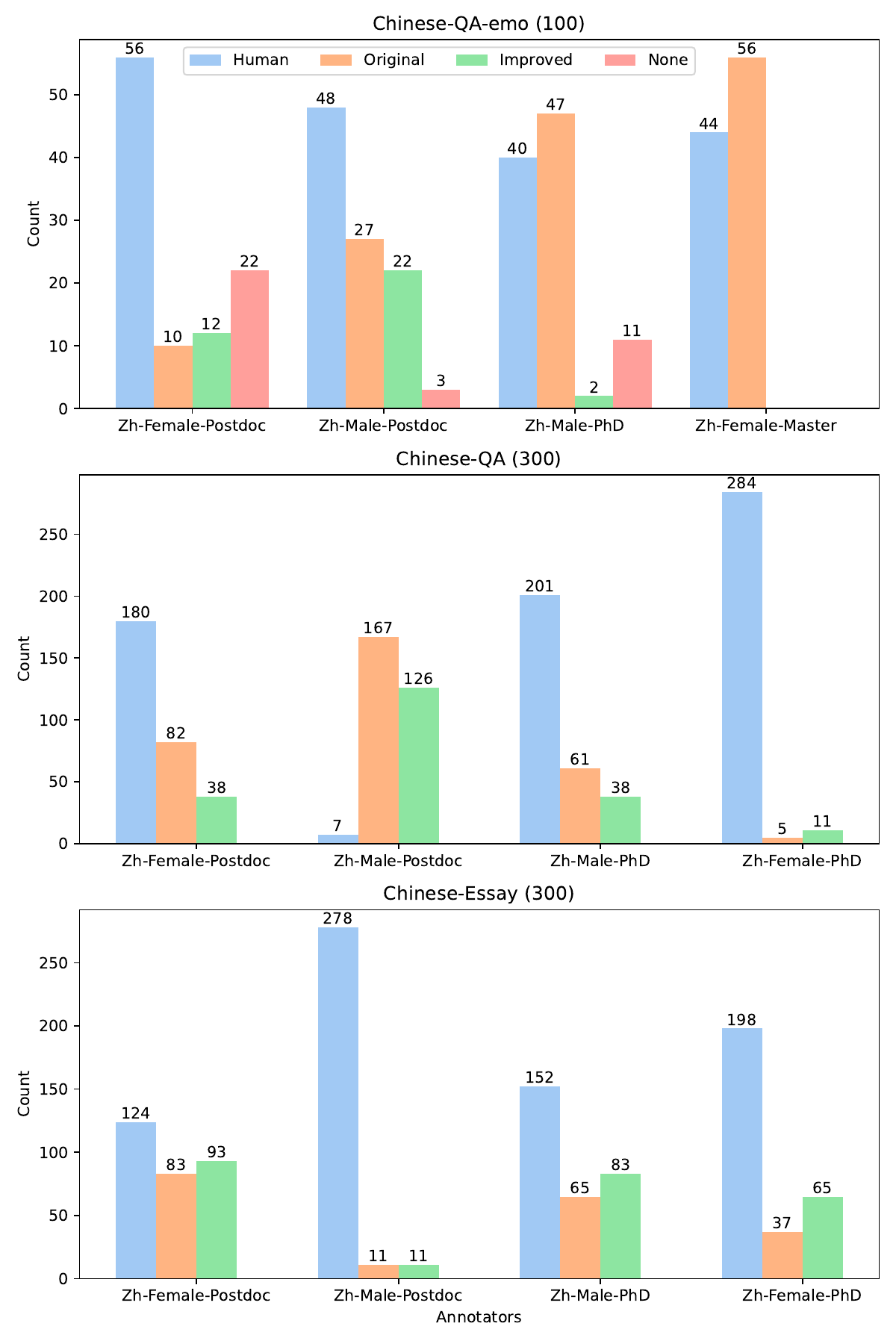}
    \caption{Human preferences for three Chinese datasets (five annotators): QA-emo is an emotion-rich question subset of Zhihu-QA with 100 examples.}
    \label{fig:pre-zh}
\end{figure}


\begin{figure}[t!]
    \centering
    \includegraphics[scale=0.45]{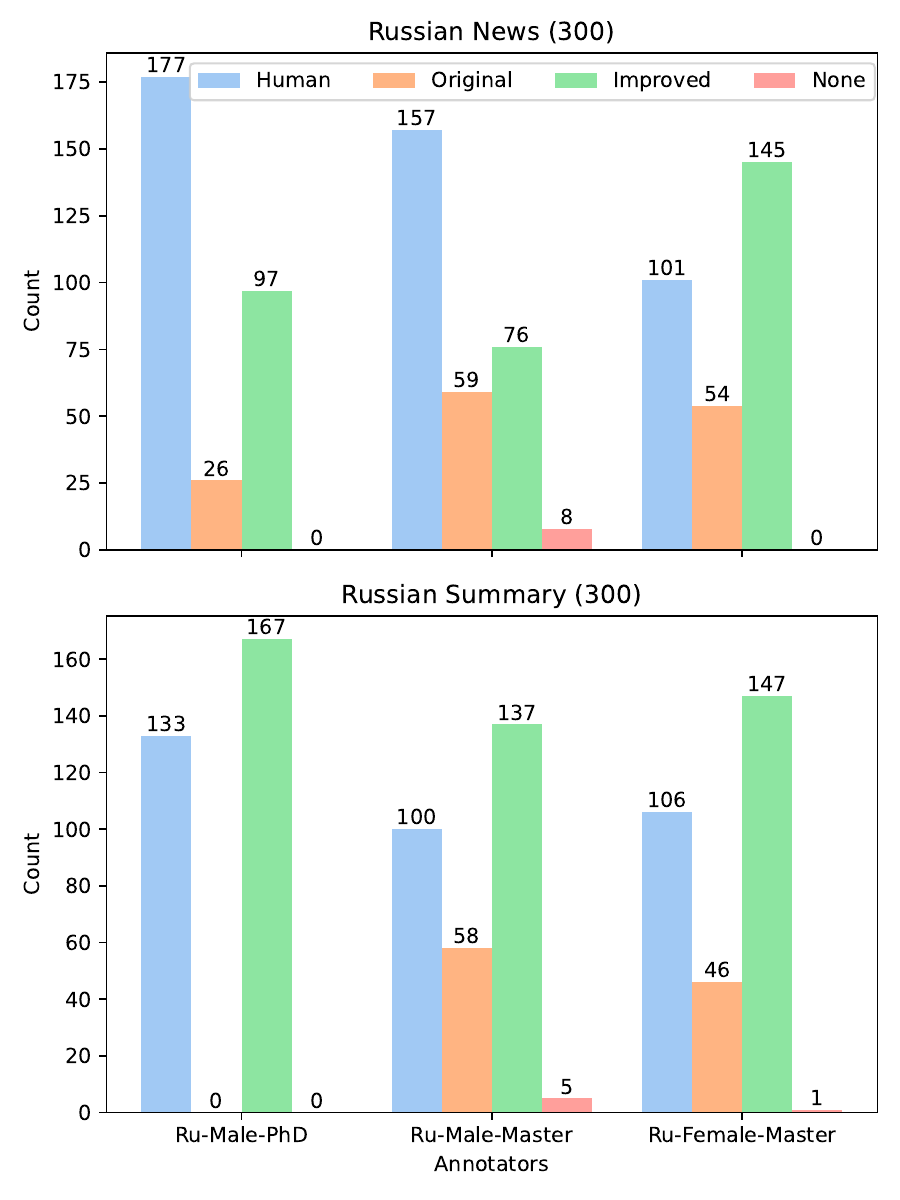}
    \includegraphics[scale=0.4]{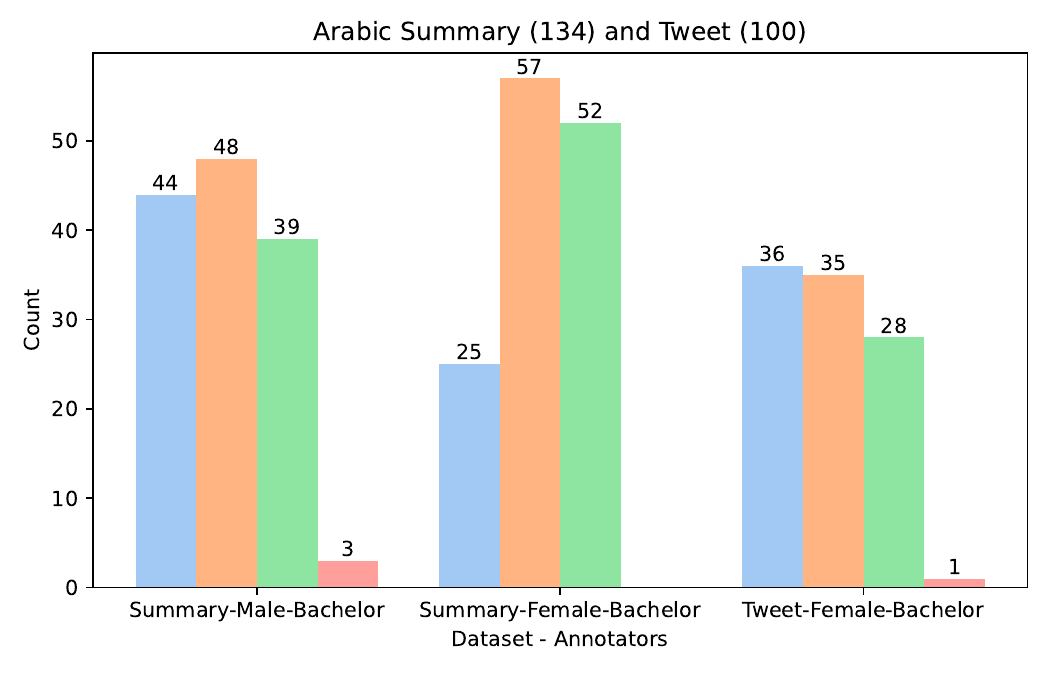}
    \caption{Human preferences for two Russian (three annotators) and two Arabic datasets (two annotators).}
    \label{fig:pre-ru-ar}
\end{figure}

\paragraph{Preference Labeling Setup:}
To evaluate the impact of human detection accuracy on their preferences, we selected datasets with high accuracy (Chinese QA and essays, Russian news), medium accuracy (Russian and Arabic summaries), and low accuracy (Arabic tweets).
For Chinese, we annotated Zhihu-QA and student essays (300 examples for each), along with 100 responses particularly for Zhihu questions.
Five unique annotators participated, identified by \textit{nationality-gender-degree}. For example, \textit{Zh-Male-PhD} refers to a Chinese male PhD student.  
We labeled Russian with three annotators and two annotators for Arabic.

\paragraph{Do People Always Prefer Human Text?}
The answer is \emph{No}.
Analyzing the preferences of ten annotators across six datasets in \figref{fig:pre-zh} and \ref{fig:pre-ru-ar}, human text was preferred in about half of the cases. Notably, for Russian and Arabic, annotators tended to favor machine text.
This is particularly evident for Russian summaries using the improved prompts (green bars) and Arabic summaries using the original prompts (orange bars).

\textit{Annotators prefer human-written text when they can clearly identify its source, otherwise they tend to favor machine text.} 
For Chinese Zhihu QA and student essays, where detection accuracy approaches 100\%, human-written text is generally preferred though there are exceptions. \textit{Zh-Male-postdoc} exhibits a unique preference distribution that deviates from the rest. The tendency to prefer human-written texts when recognized holds across languages.

Highly accurate Russian annotators prefer human-writtne texts, \textit{Ru-Male-PhD} achieved 100\% accuracy on Russian news and favored human-written text, while the other two annotators, with only medium accuracy, preferred machine-generated text. For datasets with medium or low detection accuracy (Russian and Arabic summaries, Arabic tweets), machine text was preferred in $\geq\frac{2}{3}$ of the cases. These suggest that when annotators can clearly identify human-written content, social or psychological bias toward peer-authored text emerges, in contrast, when uncertain, their intuitive preferences tend to favor machine outputs.


\paragraph{Are Humans Good Emotional Support?}
\emph{Not necessary.}
Interestingly, for emotion-rich questions (QA-emo), where human responses are expected to be more empathetic and be preferred, three out of four annotators actually preferred the machine-generated responses. The remaining annotator disliked both human and machine answers in 22\% of the cases. The annotator feedback suggested that this was influenced by the presence of mean-spirited responses from Zhihu users, where some human answers expressed personal biases and lacked empathy (see \appref{app:humanpreference}).

\paragraph{Why are Human-Written Essays Favored?}

Human-written essays tend to exhibit greater coherence and sincerity. In contrast, machine-generated essays often lack cohesion, featuring abrupt transitions and sometimes repeating the title in an attempt to simulate continuity. While LLMs can generate related concepts, they struggle to construct a compelling narrative or to seamlessly connect multiple stories under a unified theme. As a result, their outputs often feel shallow, relying on abstract concepts and rhetorical flourishes without offering substantive insight. The text tends to be verbose yet superficial, giving an impression of grandiosity without meaning. Moreover, LLM-generated essays often adopt an instructive tone, resembling answers to questions rather than peer-level discourse.

From a literary perspective, human-written articles often charm with playfulness, sincerity, or novel insights, using precise language to evoke emotion and poetic depth.
In contrast, a piece generated by an LLM lacks this literary nuance, leaving little room for contemplation or lingering thoughts. The expression is immediately clear, requiring no further reflection, resembling more of a lecture than an engaging discourse.

\paragraph{Human-Like or Liked-by-Human?}
Human texts are not always preferred. This inspires us to reflect the ultimate goal of building LLMs that are human-like vs.\ liked-by-human.
The goal of being human-like has a single target, i.e.,\ mimicking human behavior, while to be liked-by-humans involves optimization towards billions of local optima, each shaped by individual preferences.  
Current language models establish a foundation by learning from human data towards being human-like. 
As they get more advanced, they can further adapt by incorporating personal data, thus transitioning from merely imitating human behavior to aligning with individual user preferences, and moving from human-like to liked-by-human.
Our preference annotations can serve as a valuable resource for guiding models to match individual preferences in multilingual contexts.
Also, the data can facilitate investigation of the relationship between annotator characteristics (e.g.,\ MBTI personality, gender, and age) and preferences.

\section{Conclusion and Future Work}

We conducted a comprehensive study to investigate the upper bound capability of humans to detect text generated by SOTA LLMs.
Based on 16 datasets spanning 9 languages, 9 domains, and 11 LLMs, 19 expert annotators' accuracy ranged in 50-100\%, with an average of 87.6\%, revealing that it is not that challenging for experts to identify MGTs.

We found that the major gaps between human-written and machine-generated text lie in concreteness, cultural nuances, diversity in length, structure, style, and sentiment, formatting and mixture of languages in the generations. 
Prompting by explicitly indicating the distinctions could partially bridge this gap, but cultural nuances and diversity remained a challenge.
Yet, humans favored machine-generated text in over half of the cases, particularly when they could not recognize which text was written by a human.

In future work, we plan to explore human detection accuracy and preference across diverse population groups, especially their relationship with individual characteristics. How do an individual's philosophy, personality, and experience influence the detection ability and preferences. 




\section*{Acknowledgments}
We thank Ye Bai and Minghan Wang from Monash University for their valuable discussion/annotation.

\section*{Limitations}
\paragraph{Annotator Diversity} 
All annotators involved in the detection process are advanced NLP researchers, including MSc. and PhD. students, as well as postdocs, specializing in LLMs. No laymen participated in the annotation process, which ensures that the findings and the conclusions were relevant within the expert domain. It is possible that different results could be obtained if laypersons were involved in the detection tasks. However, given our aim to explore the upper bound of human capability in detecting machine-generated text, the current setting is appropriate. Future work could consider involving a broader range of participants to assess the average detection capability across different population groups.

\paragraph{Automatic Linguistic Analysis}
We did not include automated linguistic analysis using tools and techniques in this study. The reason lies in that our primary focus is on human-centered case study: how humans perceive quality of the output, identify distinguishable signals between human and machine, detection capabilities and preferences. 
There have been many studies exploring automatic MGT detection and performing linguistic analysis by automatic methods, while our aim is to understand human perception. \appref{sec:distinctionfactor} provides detailed linguistic differences (distinguishable signals) between human-written and machine-generated text which were manually summarized by expert human annotators. They are valuable insights to guide the future LLM improvement towards more human-like text generation as perceived by readers.

Our future work may explore more differences between human linguistics perceptions and insights derived from advanced computational analysis, across the three text types (human, original and improved machine-generated). 
Analysis would involve vocabulary features, sentiment analysis, language model perplexity, and other relevant metrics.

\paragraph{Statistical Significance}
In our study, each dataset includes at least 300 examples for annotation. For some datasets, 3-5 annotators labeled the same set of data, and the average values were calculated. However, a larger sample size would be beneficial to ensure more reliable results by involving multiple annotators.
In particular, the survey explores whether improved prompts bridge the gap between human-written text and machine-generated text.

We observed substantial variability between individuals. This subjectivity should be carefully considered in future work. Fortunately, three evaluation methods were used to assess whether prompting could mitigate this gap in our work, partially addressing this limitation by comparing human and automatic detection accuracy before and after the application of the improved prompts.

\section*{Ethical Statement}

\paragraph{Data Collection and Licenses}
A primary ethical consideration is the data license. We reused existing datasets for our research, such as HC3 \cite{guo-etal-2023-hc3}, M4GT-Bench \cite{wang2024m4gtbench}, OUTFOX \cite{Koike:OUTFOX:2024}, and MAGE \cite{li-etal-2024-mage}, which are publicly released under clear licensing agreements. We adhere to the intended usage of these dataset licenses.

\paragraph{Security Implications}
The datasets support the development of robust MGT detection systems, which are important for addressing security and ethical concerns. These systems help counter misinformation, reduce financial fraud, and protect content integrity in fields such as journalism, academia, and law. They also promote digital literacy by raising awareness of LLM limitations and encouraging more critical engagement with online content.

In multilingual settings, MGT detection is particularly difficult due to linguistic and cultural differences. Effective systems must address these complexities to limit disinformation, particularly in less-resourced languages. Stronger multilingual detection can improve trust in AI and reduce security risks linked to misuse.

\paragraph{Human Subject Considerations}
Our study involved evaluators distinguishing between human-written and MGT and expressing their preferences. All annotators gave informed consent, understood the study's objectives, and could withdraw at any time. Because the study focused on expert evaluation, the findings may not generalize to laypeople. Future work should include more diverse participants across demographic and professional groups.

\paragraph{Transparency and Reproducibility}
With the aim to promote open research, we release our collected datasets to the public, including all human annotations and preference rankings, enabling other researchers to build upon our work. We further provide comprehensive documentation to ensure full reproducibility and transparency.

\bibliography{ref}

\begin{thebibliography}{35}
\expandafter\ifx\csname natexlab\endcsname\relax\def\natexlab#1{#1}\fi

\bibitem[{Anthropic(2024)}]{Claude3}
Anthropic. 2024.
\newblock \href {https://api.semanticscholar.org/CorpusID:268232499} {The {C}laude 3 model family: {O}pus, {S}onnet, {H}aiku}.

\bibitem[{Bonisoli et~al.(2023)Bonisoli, di~Buono, Po, and Rollo}]{bonisoli2023}
Giovanni Bonisoli, Maria~Pia di~Buono, Laura Po, and Federica Rollo. 2023.
\newblock \href {https://doi.org/10.1145/3539618.3591904} {{DICE}: A dataset of {I}talian crime event news}.
\newblock In \emph{Proceedings of the 46th International ACM SIGIR Conference on Research and Development in Information Retrieval}, SIGIR~'23, Taipei, Taiwan. {ACM}.

\bibitem[{Chein et~al.(2024)Chein, Martinez, and Barone}]{chein2024human}
Jason~M. Chein, Steven~A. Martinez, and Alexander~R. Barone. 2024.
\newblock \href {https://doi.org/10.1038/s41598-024-76218-y} {Human intelligence can safeguard against artificial intelligence: Individual differences in the discernment of human from {A}{I} texts}.
\newblock \emph{Scientific Reports}, 14(1):25989.

\bibitem[{Clark et~al.(2021)Clark, August, Serrano, Haduong, Gururangan, and Smith}]{clark-etal-2021-thats}
Elizabeth Clark, Tal August, Sofia Serrano, Nikita Haduong, Suchin Gururangan, and Noah~A. Smith. 2021.
\newblock \href {https://doi.org/10.18653/v1/2021.acl-long.565} {All that`s {\textquoteleft}human' is not gold: Evaluating human evaluation of generated text}.
\newblock In \emph{Proceedings of the 59th Annual Meeting of the Association for Computational Linguistics and the 11th International Joint Conference on Natural Language Processing (Volume 1: Long Papers)}, ACL-IJCNLP~'21, pages 7282--7296, Online. Association for Computational Linguistics.

\bibitem[{Dubey et~al.(2024)Dubey, Jauhri, Pandey, Kadian, Al-Dahle, Letman, Mathur, Schelten, Yang, Fan et~al.}]{llama3}
Abhimanyu Dubey, Abhinav Jauhri, Abhinav Pandey, Abhishek Kadian, Ahmad Al-Dahle, Aiesha Letman, Akhil Mathur, Alan Schelten, Amy Yang, Angela Fan, et~al. 2024.
\newblock \href {http://arxiv.org/abs/2407.21783} {The {L}lama 3 herd of models}.
\newblock \emph{ArXiv preprint}, arXiv:2407.21783.

\bibitem[{Dugan et~al.(2023)Dugan, Ippolito, Kirubarajan, Shi, and Callison-Burch}]{liam2023reakfake}
Liam Dugan, Daphne Ippolito, Arun Kirubarajan, Sherry Shi, and Chris Callison-Burch. 2023.
\newblock \href {https://ojs.aaai.org/index.php/AAAI/article/view/26501} {Real or fake text?: Investigating human ability to detect boundaries between human-written and machine-generated text}.
\newblock In \emph{Proceedings of the 37th AAAI Conference on Artificial Intelligence}, AAAI~'23, pages 12763--12771, Washington, DC, USA.

\bibitem[{Einea et~al.(2019)Einea, Elnagar, and {Al Debsi}}]{EINEA2019104076}
Omar Einea, Ashraf Elnagar, and Ridhwan {Al Debsi}. 2019.
\newblock \href {https://doi.org/https://doi.org/10.1016/j.dib.2019.104076} {{S}{A}{N}{A}{D}: Single-label {A}rabic news articles dataset for automatic text categorization}.
\newblock \emph{Data in Brief}, 25:104076.

\bibitem[{Elangovan et~al.(2024)Elangovan, Liu, Xu, Bodapati, and Roth}]{elangovan-etal-2024-considers}
Aparna Elangovan, Ling Liu, Lei Xu, Sravan~Babu Bodapati, and Dan Roth. 2024.
\newblock \href {https://doi.org/10.18653/v1/2024.acl-long.63} {{C}on{S}i{DERS}-the-human evaluation framework: Rethinking human evaluation for generative large language models}.
\newblock In \emph{Proceedings of the 62nd Annual Meeting of the Association for Computational Linguistics (Volume 1: Long Papers)}, ACL~'24, pages 1137--1160, Bangkok, Thailand. Association for Computational Linguistics.

\bibitem[{Garbacea et~al.(2019)Garbacea, Carton, Yan, and Mei}]{garbacea-etal-2019-judge}
Cristina Garbacea, Samuel Carton, Shiyan Yan, and Qiaozhu Mei. 2019.
\newblock \href {https://doi.org/10.18653/v1/D19-1409} {Judge the judges: A large-scale evaluation study of neural language models for online review generation}.
\newblock In \emph{Proceedings of the 2019 Conference on Empirical Methods in Natural Language Processing and the 9th International Joint Conference on Natural Language Processing}, EMNLP-IJCNLP~'19, pages 3968--3981, Hong Kong, China. Association for Computational Linguistics.

\bibitem[{Guo et~al.(2023)Guo, Zhang, Wang, Jiang, Nie, Ding, Yue, and Wu}]{guo-etal-2023-hc3}
Biyang Guo, Xin Zhang, Ziyuan Wang, Minqi Jiang, Jinran Nie, Yuxuan Ding, Jianwei Yue, and Yupeng Wu. 2023.
\newblock \href {https://arxiv.org/abs/2301.07597} {How close is {C}hat{G}{P}{T} to human experts? comparison corpus, evaluation, and detection}.
\newblock \emph{ArXiv preprint}, arXiv:2301.07597.

\bibitem[{Hasan et~al.(2021)Hasan, Bhattacharjee, Islam, Mubasshir, Li, Kang, Rahman, and Shahriyar}]{hasan-etal-2021-xl}
Tahmid Hasan, Abhik Bhattacharjee, Md.~Saiful Islam, Kazi Mubasshir, Yuan-Fang Li, Yong-Bin Kang, M.~Sohel Rahman, and Rifat Shahriyar. 2021.
\newblock \href {https://doi.org/10.18653/v1/2021.findings-acl.413} {{X}{L}-{S}um: Large-scale multilingual abstractive summarization for 44 languages}.
\newblock In \emph{Findings of the Association for Computational Linguistics}, ACL-IJCNLP~'21, pages 4693--4703, Online. Association for Computational Linguistics.

\bibitem[{Hitsuwari et~al.(2023)Hitsuwari, Ueda, Yun, and Nomura}]{jimpei2023creative}
Jimpei Hitsuwari, Yoshiyuki Ueda, Woojin Yun, and Michio Nomura. 2023.
\newblock \href {https://doi.org/10.1016/J.CHB.2022.107502} {Does human-{A}{I} collaboration lead to more creative art? {A}esthetic evaluation of human-made and ai-generated haiku poetry}.
\newblock \emph{Computers in Human Behavior}, 139:107502.

\bibitem[{Ippolito et~al.(2020)Ippolito, Duckworth, Callison-Burch, and Eck}]{ippolito-etal-2020-automatic}
Daphne Ippolito, Daniel Duckworth, Chris Callison-Burch, and Douglas Eck. 2020.
\newblock \href {https://doi.org/10.18653/v1/2020.acl-main.164} {Automatic detection of generated text is easiest when humans are fooled}.
\newblock In \emph{Proceedings of the 58th Annual Meeting of the Association for Computational Linguistics}, ACL~'20, pages 1808--1822, Online. Association for Computational Linguistics.

\bibitem[{Koike et~al.(2024)Koike, Kaneko, and Okazaki}]{Koike:OUTFOX:2024}
Ryuto Koike, Masahiro Kaneko, and Naoaki Okazaki. 2024.
\newblock \href {https://doi.org/10.1609/AAAI.V38I19.30120} {{OUTFOX:} {LLM}-generated essay detection through in-context learning with adversarially generated examples}.
\newblock In \emph{Proceedings of the 38th {AAAI} Conference on Artificial Intelligence}, AAAI~'24, pages 21258--21266, Vancouver, Canada. {AAAI} Press.

\bibitem[{Li et~al.(2023)Li, Hovy, and Lau}]{peersum_2023}
Miao Li, Eduard Hovy, and Jey~Han Lau. 2023.
\newblock \href {https://doi.org/10.18653/v1/2023.findings-emnlp.472} {Summarizing multiple documents with conversational structure for meta-review generation}.
\newblock In \emph{Findings of the Association for Computational Linguistics}, EMNLP~'23, pages 7089--7112, Singapore. Association for Computational Linguistics.

\bibitem[{Li et~al.(2024{\natexlab{a}})Li, Li, Cui, Bi, Wang, Wang, Yang, Shi, and Zhang}]{li-etal-2024-mage}
Yafu Li, Qintong Li, Leyang Cui, Wei Bi, Zhilin Wang, Longyue Wang, Linyi Yang, Shuming Shi, and Yue Zhang. 2024{\natexlab{a}}.
\newblock \href {https://doi.org/10.18653/v1/2024.acl-long.3} {{MAGE}: {M}achine-generated text detection in the wild}.
\newblock In \emph{Proceedings of the 62nd Annual Meeting of the Association for Computational Linguistics (Volume 1: Long Papers)}, ACL~'24, pages 36--53, Bangkok, Thailand. Association for Computational Linguistics.

\bibitem[{Li et~al.(2024{\natexlab{b}})Li, Chen, Shao, Xie, Jiang, and Nie}]{li2024ecot}
Zaijing Li, Gongwei Chen, Rui Shao, Yuquan Xie, Dongmei Jiang, and Liqiang Nie. 2024{\natexlab{b}}.
\newblock \href {https://arxiv.org/abs/2401.06836} {Enhancing emotional generation capability of large language models via emotional chain-of-thought}.
\newblock \emph{ArXiv preprint}, arXiv:2401.06836.

\bibitem[{Markowitz et~al.(2024)Markowitz, Hancock, and Bailenson}]{markowitz2024linguistic}
David~M Markowitz, Jeffrey~T Hancock, and Jeremy~N Bailenson. 2024.
\newblock \href {https://doi.org/10.1177/0261927X231200201} {Linguistic markers of inherently false {A}{I} communication and intentionally false human communication: {E}vidence from hotel reviews}.
\newblock \emph{Journal of Language and Social Psychology}, 43(1):63--82.

\bibitem[{Mieczkowski et~al.(2021)Mieczkowski, Hancock, Naaman, Jung, and Hohenstein}]{mieczkowski2021ai}
Hannah Mieczkowski, Jeffrey~T. Hancock, Mor Naaman, Malte Jung, and Jess Hohenstein. 2021.
\newblock \href {https://doi.org/10.1145/3449091} {{AI}-mediated communication: Language use and interpersonal effects in a referential communication task}.
\newblock \emph{Proceedings of the ACM on Human-Computer Interaction}, 5(CSCW1):1--14.

\bibitem[{{MOP-LIWU Community} and {MNBVC Team}(2023)}]{mnbvc}
{MOP-LIWU Community} and {MNBVC Team}. 2023.
\newblock {MNBVC}: Massive never-ending {BT} vast {C}hinese corpus.
\newblock \url{https://github.com/esbatmop/MNBVC}.

\bibitem[{OpenAI(2023)}]{OpenAI2023GPT4TR}
OpenAI. 2023.
\newblock \href {https://api.semanticscholar.org/CorpusID:257532815} {{G}{P}{T}-4 technical report}.
\newblock \emph{ArXiv preprint}, arXiv:2303.08774.

\bibitem[{Pedrotti et~al.(2025)Pedrotti, Papucci, Ciaccio, Miaschi, Puccetti, Dell{'}Orletta, and Esuli}]{pedrotti-etal-2025-stress}
Andrea Pedrotti, Michele Papucci, Cristiano Ciaccio, Alessio Miaschi, Giovanni Puccetti, Felice Dell{'}Orletta, and Andrea Esuli. 2025.
\newblock \href {https://doi.org/10.18653/v1/2025.findings-acl.156} {Stress-testing machine generated text detection: Shifting language models writing style to fool detectors}.
\newblock In \emph{Findings of the Association for Computational Linguistics}, ACL~'25, pages 3010--3031, Vienna, Austria. Association for Computational Linguistics.

\bibitem[{Polignano et~al.(2024)Polignano, Basile, and Semeraro}]{polignano2024advanced}
Marco Polignano, Pierpaolo Basile, and Giovanni Semeraro. 2024.
\newblock \href {http://arxiv.org/abs/2405.07101} {Advanced natural-based interaction for the {Italian} language: {LLaMAntino}-3-{ANITA}}.
\newblock \emph{ArXiv preprint}, arXiv:2405.07101.

\bibitem[{Russell et~al.(2025)Russell, Karpinska, and Iyyer}]{russell-etal-2025-people}
Jenna Russell, Marzena Karpinska, and Mohit Iyyer. 2025.
\newblock \href {https://doi.org/10.18653/v1/2025.acl-long.267} {People who frequently use {C}hat{GPT} for writing tasks are accurate and robust detectors of {AI}-generated text}.
\newblock In \emph{Proceedings of the 63rd Annual Meeting of the Association for Computational Linguistics (Volume 1: Long Papers)}, ACL~'25, pages 5342--5373, Vienna, Austria. Association for Computational Linguistics.

\bibitem[{Shamardina et~al.(2022)Shamardina, Mikhailov, Chernianskii, Fenogenova, Saidov, Valeeva, Shavrina, Smurov, Tutubalina, and Artemova}]{shamardina2022findings}
Tatiana Shamardina, Vladislav Mikhailov, Daniil Chernianskii, Alena Fenogenova, Marat Saidov, Anastasiya Valeeva, Tatiana Shavrina, Ivan Smurov, Elena Tutubalina, and Ekaterina Artemova. 2022.
\newblock \href {http://arxiv.org/abs/arXiv:2206.01583} {Findings of the the {R}u{A}{T}{D} shared task 2022 on artificial text detection in {R}ussian}.
\newblock \emph{ArXiv preprint}, arXiv:2206.01583.

\bibitem[{Shamardina et~al.(2025)Shamardina, Saidov, Fenogenova, Tumanov, Zemlyakova, Lebedeva, Gryaznova, Shavrina, Mikhailov, and Artemova}]{shamardina2025coat}
Tatiana Shamardina, Marat Saidov, Alena Fenogenova, Aleksandr Tumanov, Alina Zemlyakova, Anna Lebedeva, Ekaterina Gryaznova, Tatiana Shavrina, Vladislav Mikhailov, and Ekaterina Artemova. 2025.
\newblock \href {https://doi.org/10.1017/nlp.2024.38} {{C}o{A}{T}: Corpus of artificial texts}.
\newblock \emph{Natural Language Processing}, 31(1):150--175.

\bibitem[{Song et~al.(2020)Song, Zhang, Fu, Liu, Liu, and Cheng}]{song-etal-2020-multi}
Wei Song, Kai Zhang, Ruiji Fu, Lizhen Liu, Ting Liu, and Miaomiao Cheng. 2020.
\newblock \href {https://doi.org/10.18653/v1/2020.emnlp-main.546} {Multi-stage re-training for automated {C}hinese essay scoring}.
\newblock In \emph{Proceedings of the 2020 Conference on Empirical Methods in Natural Language Processing}, EMNLP~'20, pages 6723--6733, Online. Association for Computational Linguistics.

\bibitem[{Su et~al.(2025)Su, Wang, Wan, Zhang, and Luo}]{su-etal-2025-haco}
Zhixiong Su, Yichen Wang, Herun Wan, Zhaohan Zhang, and Minnan Luo. 2025.
\newblock \href {https://doi.org/10.18653/v1/2025.acl-long.1069} {{HAC}o-det: A study towards fine-grained machine-generated text detection under human-{AI} coauthoring}.
\newblock In \emph{Proceedings of the 63rd Annual Meeting of the Association for Computational Linguistics (Volume 1: Long Papers)}, ACL~'25, pages 22015--22036, Vienna, Austria. Association for Computational Linguistics.

\bibitem[{Ta et~al.(2026)Ta, Van, Hoang, Le-Anh, Nguyen, Nguyen, Wang, Nakov, and Sang}]{ta-etal-2026-faid}
Minh~Ngoc Ta, Dong~Cao Van, Duc-Anh Hoang, Minh Le-Anh, Truong Nguyen, My~Anh~Tran Nguyen, Yuxia Wang, Preslav Nakov, and Dinh~Viet Sang. 2026.
\newblock \href {https://doi.org/10.18653/v1/2026.eacl-long.151} {{F}{A}{I}{D}: {F}ine-grained {A}{I}-generated {T}ext {D}etection using {M}ulti-task {A}uxiliary and {M}ulti-level {C}ontrastive {L}earning}.
\newblock In \emph{Proceedings of the 19th Conference of the {E}uropean Chapter of the {A}ssociation for {C}omputational {L}inguistics (Volume 1: Long Papers)}, EACL~'26, pages 3275--3296, Rabat, Morocco. Association for Computational Linguistics.

\bibitem[{Team et~al.(2023)Team, Anil, Borgeaud, Alayrac, Yu, Soricut, Schalkwyk, Dai, Hauth, Millican et~al.}]{team2023gemini}
Gemini Team, Rohan Anil, Sebastian Borgeaud, Jean-Baptiste Alayrac, Jiahui Yu, Radu Soricut, Johan Schalkwyk, Andrew~M Dai, Anja Hauth, Katie Millican, et~al. 2023.
\newblock \href {https://arxiv.org/abs/2312.11805} {Gemini: a family of highly capable multimodal models}.
\newblock \emph{ArXiv preprint}, arXiv:2312.11805.

\bibitem[{van~der Lee et~al.(2019)van~der Lee, Gatt, van Miltenburg, Wubben, and Krahmer}]{van-der-lee-etal-2019-best}
Chris van~der Lee, Albert Gatt, Emiel van Miltenburg, Sander Wubben, and Emiel Krahmer. 2019.
\newblock \href {https://doi.org/10.18653/v1/W19-8643} {Best practices for the human evaluation of automatically generated text}.
\newblock In \emph{Proceedings of the 12th International Conference on Natural Language Generation}, INLG~'19, pages 355--368, Tokyo, Japan. Association for Computational Linguistics.

\bibitem[{Wang et~al.(2024{\natexlab{a}})Wang, Mansurov, Ivanov, Su, Shelmanov, Tsvigun, Mohammed~Afzal, Mahmoud, Puccetti, Arnold, Aji, Habash, Gurevych, and Nakov}]{wang2024m4gtbench}
Yuxia Wang, Jonibek Mansurov, Petar Ivanov, Jinyan Su, Artem Shelmanov, Akim Tsvigun, Osama Mohammed~Afzal, Tarek Mahmoud, Giovanni Puccetti, Thomas Arnold, Alham Aji, Nizar Habash, Iryna Gurevych, and Preslav Nakov. 2024{\natexlab{a}}.
\newblock \href {https://doi.org/10.18653/v1/2024.acl-long.218} {{M}4{G}{T}-{B}ench: Evaluation benchmark for black-box machine-generated text detection}.
\newblock In \emph{Proceedings of the 62nd Annual Meeting of the Association for Computational Linguistics (Volume 1: Long Papers)}, ACL~'24, pages 3964--3992, Bangkok, Thailand. Association for Computational Linguistics.

\bibitem[{Wang et~al.(2024{\natexlab{b}})Wang, Mansurov, Ivanov, Su, Shelmanov, Tsvigun, Whitehouse, Mohammed~Afzal, Mahmoud, Sasaki, Arnold, Aji, Habash, Gurevych, and Nakov}]{wang-etal-2024-m4}
Yuxia Wang, Jonibek Mansurov, Petar Ivanov, Jinyan Su, Artem Shelmanov, Akim Tsvigun, Chenxi Whitehouse, Osama Mohammed~Afzal, Tarek Mahmoud, Toru Sasaki, Thomas Arnold, Alham Aji, Nizar Habash, Iryna Gurevych, and Preslav Nakov. 2024{\natexlab{b}}.
\newblock \href {https://aclanthology.org/2024.eacl-long.83} {{M}4: Multi-generator, multi-domain, and multi-lingual black-box machine-generated text detection}.
\newblock In \emph{Proceedings of the 18th Conference of the European Chapter of the Association for Computational Linguistics (Volume 1: Long Papers)}, EACL~'24, pages 1369--1407, St. Julian{'}s, Malta. Association for Computational Linguistics.

\bibitem[{Wang et~al.(2025)Wang, Shelmanov, Mansurov, Tsvigun, Mikhailov, Xing, Xie, Geng, Puccetti, Artemova, Su, Ta, Abassy, Elozeiri, El~Etter, Goloburda, Mahmoud, Tomar, Laiyk, Mohammed~Afzal, Koike, Kaneko, Aji, Habash, Gurevych, and Nakov}]{wang-etal-2025-genai}
Yuxia Wang, Artem Shelmanov, Jonibek Mansurov, Akim Tsvigun, Vladislav Mikhailov, Rui Xing, Zhuohan Xie, Jiahui Geng, Giovanni Puccetti, Ekaterina Artemova, Jinyan Su, Minh~Ngoc Ta, Mervat Abassy, Kareem~Ashraf Elozeiri, Saad El Dine~Ahmed El~Etter, Maiya Goloburda, Tarek Mahmoud, Raj~Vardhan Tomar, Nurkhan Laiyk, Osama Mohammed~Afzal, Ryuto Koike, Masahiro Kaneko, Alham~Fikri Aji, Nizar Habash, Iryna Gurevych, and Preslav Nakov. 2025.
\newblock \href {https://aclanthology.org/2025.genaidetect-1.27/} {{G}en{A}{I} content detection task 1: English and multilingual machine-generated text detection: {A}{I} vs. human}.
\newblock In \emph{Proceedings of the 1st Workshop on {GenAI} Content Detection}, GenAIDetect~'25, pages 244--261, Abu Dhabi, UAE. International Conference on Computational Linguistics.

\bibitem[{Yeshpanov et~al.(2024)Yeshpanov, Efimov, Boytsov, Shalkarbayuli, and Braslavski}]{kazqad}
Rustem Yeshpanov, Pavel Efimov, Leonid Boytsov, Ardak Shalkarbayuli, and Pavel Braslavski. 2024.
\newblock \href {https://aclanthology.org/2024.lrec-main.843/} {{K}az{QAD}: {K}azakh open-domain question answering dataset}.
\newblock In \emph{Proceedings of the 2024 Joint International Conference on Computational Linguistics, Language Resources and Evaluation}, LREC-COLING~'24, pages 9645--9656, Torino, Italia. ELRA and ICCL.

\end{thebibliography}
\bibliographystyle{acl_natbib}

\clearpage
\section*{Appendix}
\appendix

\section{Related Work}
\label{sec:relatedwork}
We first review studies investigating human evaluation of detecting MGT, and then summarize distinguishable features between human-written and machine-generated text, and end up with a discussion of relationship between individual factors, human detection capability and preference.

\subsection{Human Detection on MGT}
\label{sec:human-eval-MGT}
Early studies probing the outputs of less advanced generative models found that human evaluators could detect the differences.
For example, \citet{garbacea-etal-2019-judge} reported that individual evaluators obtained 66.61\% accuracy when evaluating online product reviews generated by 12 sequential models before 2019, and majority voting of five annotators' predictions improved to 72.63\%. \citet{ippolito-etal-2020-automatic} found that trained evaluators were able to detect GPT2-large text 71.4\%.


However, since the emergence of GPT-3, it has become challenging for human evaluators to distinguish between human-written and machine-generated text. Evaluators could guess GPT-3-generated stories, news articles and recipes with 50\% accuracy~\citep{clark-etal-2021-thats}. 203 crowd-soured workers employed from Prolific show 57\% average accuracy when discerning news generated by GPT-3.5-turbo and news collected from trusted news organizations, e.g., New York Times, Wall Street Journal, and they obtained 78\% on social media comments, both with a wide spread in performance across individuals~\citep{chein2024human}. \citet{guo-etal-2023-hc3} also showed large differences in detection capability between experts and laymen, where the former is much higher than the later across different domains. Average accuracy of experts is above 90\%, but random guess for amateurs, i.e., 48\% and 54\% for English and Chinese respectively. With expert annotators, academic paper abstracts are harder to detect than the Reddit answers. Detection accuracy ranges from 41 to 94 for Reddit with an average of 77, while 72 for abstracts with individual performance ranging from 60 to 84, revealing that scientific materials are more challenging even for experts who are researchers and frequent users of LLMs~\cite{wang-etal-2024-m4}. This variability highlights substantial uncertainty in human judgment and suggests the need for better training, calibration protocols, and complementary automatic detection tools in evaluation pipelines.


In addition to binary human vs. AI detection, \citet{wang2024m4gtbench} performed 4-class detection to identify which LLM generated a text, achieving 21.2\% accuracy, below the 25\% random baseline, despite NLP PhD annotators. \citet{liam2023reakfake} asked annotators to select boundary sentences between human- and machine-authored snippets in mixed collaborative text, finding 23.4\% accuracy (chance 10\%).

As shown in \tabref{tab:human-mgt-detection-survey}, previous studies primarily focus on English and outputs of GPT-3.5-turbo, non-English languages and more advanced models have been under-investigated. The accuracy of human detection is influenced by languages, generative models, and domains of the given text, as well as the annotators. Large variance is observed among individual evaluators, particularly between experts and laymen.
Therefore, we aim to fill this gap by a comprehensive case study exploring whether human can safeguard against AI over 16 dataset spanning nine languages, nine domains and 11 SOTA LLMs in four evaluation setups.

\subsection{Distinguishable Signals} 
To characterize the specific linguistic properties that distinguish human from AI texts, \citet{markowitz2024linguistic} compared real human hotel reviews to a set of LLM-created fake hotel reviews, and observed that machine texts had a more ``analytical'' style and exhibited increased use of affective language (stronger positive ``emotional tone''). Similar work has shown that AI ``smart replies'' likewise demonstrate an emotional positivity bias~\cite{mieczkowski2021ai}.  \citet{guo-etal-2023-hc3} summarized that GPT-3.5-turbo writes in an organized manner, with clear logic, tending to offer a long and detailed answer. Also, the model responses generally strictly focused on the given question, whereas humans’ are divergent and easily shift to other topics. Model answers are objective, typically formal, while humans’ are more subjective, colloquial and emotion-rich. \citet{wang2024m4gtbench} reported that human-written texts have some imperfect formatting patterns compared to machine text, e.g., initial double newlines, spaces, missing new lines in the paragraph, typos, inconsistencies within texts, or specific references and URLs. \citet{su-etal-2025-haco, ta-etal-2026-faid} further shows that many real-world texts are co-authored by humans and LLMs, requiring fine-grained detection beyond document-level classification.



Recent work further highlights the importance of experience with LLMs for improving detection ability. \citet{russell-etal-2025-people} show that annotators who frequently use ChatGPT for writing tasks can achieve near-perfect detection accuracy, significantly outperforming both lay annotators and existing automatic detectors, even under adversarial conditions such as paraphrasing.
Beyond probing the upper bound of human detection ability, we also investigate whether the linguistic patterns discussed above generalize across languages and more advanced AI systems, and whether they can support more accurate judgments of text origin.

\subsection{Detection Accuracy and Preference}
\label{sec:whyexperts}

\textbf{What factors influence individuals' detection accuracy?}
\citet{chein2024human} examined whether psychological attributes (e.g., fluid intelligence, executive functioning, empathy) and experience (e.g., smartphone and social media use) predict the ability to distinguish human from AI-generated text. They found that only fluid intelligence—the capacity to reason and solve novel problems without prior knowledge—was strongly associated with detection performance, while other factors showed no significant correlation.\footnote{\emph{Fluid intelligence} refers to the ability to think abstractly, recognize patterns, make decisions in unfamiliar situations, and adapt to new challenges. It is often linked to biological factors and tends to peak in early adulthood, in contrast to \emph{crystallized intelligence}, which reflects accumulated knowledge and experience.} Moreover, increased exposure to mixed human and AI content online does not improve detection ability; instead, it may make AI-generated text appear more human-like.

Overall, language proficiency, cognitive ability, and familiarity with AI outputs likely influence detection accuracy. \citet{pedrotti-etal-2025-stress} further show that detection systems can be easily deceived by rewriting strategies that shift machine-generated text toward human-like distributions, leading to substantial performance drops.

Strong language proficiency helps identify overly formal, repetitive, or mechanically structured phrasing typical of AI outputs. Native speakers with cultural awareness are better equipped to detect content lacking subtlety or creativity. Similarly, strong analytical skills aid in spotting inconsistencies, logical flaws, and stylistic patterns indicative of machine-generated text. Frequent interaction with LLMs may help individuals recognize common generation patterns, such as positivity bias~\cite{guo-etal-2023-hc3, chein2024human}.

To estimate the upper bound of human detection capability for current SOTA LLMs, all annotators in our study are native speakers with advanced expertise, including MSc., PhD, and postdoctoral researchers working on LLMs (5 female and 14 male).

\section{Annotation Tool}
To mitigate potential labeling biases arising from raw spreadsheet annotation and to enhance efficiency, we implemented two methods with optimized interfaces and workflows for our annotation: (1)~a custom pipeline using the Google Workspace suite, including Apps Script, Google Sheets, and Google Forms. The core idea was to store all data in Google Sheets, use Apps Script to extract data and generate a survey in Google Forms;\footnote{https://developers.google.com/apps-script} and (2)~Label Studio, an open-source multi-type data labeling and annotation tool with a standardized output format. We designed a custom template for our annotation task and collected results using this platform. The annotators were given the choice to use their preferred tool.\footnote{https://github.com/HumanSignal/label-studio/}

\section{Datasets}
\label{sec:datasets}

\begin{table*}[ht!]
    \centering
    \small
    \resizebox{\textwidth}{!}{
    \begin{tabular}{lllr|ccclr}
    \toprule
    \textbf{Language} & \textbf{Source/} & \textbf{Data} &  \textbf{Total} & \multicolumn{5}{c}{\textbf{Sampled Parallel Data}}  \\
                      & \textbf{Domain} & \textbf{License} &\textbf{Human} & \textbf{Human} & \textbf{GPT-4o} & \textbf{Claude} &  \textbf{LLM2 Name (\#)} & \textbf{Total}  \\
    \midrule
    \multirow{4}{*}{Arabic} & Dialect Tweet &  Apache 2.0 &  1400 & 300 & 300* & -- & Qwen2 (300*) & 900 \\
    & EASC & cc-by-sa-3.0 & 765 & 153 & 153 & -- & -- & 306 \\
    & Youm7 News & --- & 21,000 & 1,000 & 1,000 & -- & AceGPT (1,000) & 3,000 \\
    & SANAD & cc-by-4.0 & 194,797 & 100 & 100 & -- & -- & 200 \\
    \midrule
    \multirow{4}{*}{Chinese} & Zhihu-QA & cc-by-4.0 & 224,761 & 588 & 588 & -- & Qwen-turbo (588) & 1,764 \\
                             & Student essay & cc-by-4.0 & 93,002 & 600 & -- & 300* & ChatGLM4 (300*) & 1,200 \\
                             & Student essay & cc-by-4.0 & 51 & 51 & -- & 51 &  ChatGLM4 (51) & 153 \\
                             & Government Report & MIT & 200,409& 500 & 500 & -- &  Baichuan2-13B (500) & 1,500 \\
    \midrule
    English & Peersum~\citep{peersum_2023} & cc-by-sa-4.0 &  5,158 & 400 & 200 & 200  & -- &  800\\
    \midrule
    Hindi & News & cc-by-4.0 & 3,995 & 600 & 600 & -- & -- & 1,200 \\
    \midrule
    Italian & DICE & cc-by-sa &  10,518 & 300 & 300 & Anita (300) &  Llama3-405B (300) & 1,200\\
    \midrule
    Japanese & News & cc-by-nc-sa-4.0 & 7,110 & 300 & 300 & -- & -- & 600 \\
     \midrule
    Kazakh & Wikipedia & cc-by-sa-4.0 &  4,827 & 300 & 300  & -- & -- & 600\\
    \midrule
    \multirow{2}{*}{Russian} 
    & News & MIT & 800,000 & 300 & 300 & -- & Vikhr-Nemo-12B (300) & 900 \\
    & Academic summary & MIT & 31,000 & 300 & 300 & -- & Vikhr-Nemo-12B (300) & 900 \\
    \midrule
    \multirow{2}{*}{Vietnamese} 
    & Wikipedia & cc-by-sa-3.0 & 600 & 600 & 600 & -- & -- & 1,200 \\
    & News & Kaggle data & 290,282 & 600 & 600 & -- & -- & 1,200 \\
    \midrule
    \bf Total & -- & -- & 1,690,803	& 6,992 & 6,141 & 851 & 3,639 & \textbf{17,623} \\
    \bottomrule
    \end{tabular}
   }
    \caption{Statistics of multilingual data for human annotation. Machine data with * means non-parallel data. }
    \label{tab:multilingual-data}
\end{table*}

\tabref{tab:multilingual-data} gives statistics about the 16 datasets.




\subsection{Arabic}
We collected three categories of Arabic text (tweets, summaries, and news), covering four datasets.

\paragraph{Arabic Dialect Tweets}
We randomly sampled 300 tweets from the QCRI Arabic POS Dialect dataset\footnote{\url{https://huggingface.co/datasets/QCRI/arabic_pos_dialect}}
, which includes four dialects: Egyptian, Moroccan, Gulf, and Levantine. Originally developed for part-of-speech tagging, the dataset spans a wide range of topics. We preprocessed the tweets by removing usernames and normalizing the text into sentence form.

We then generated 600 machine-produced tweets (150 per dialect) using \gptfouro and Qwen-2 (7.5B). Rather than using APIs, we collected outputs via the chat interface to better observe generation patterns. The prompt used was: \textit{write a random tweet in \texttt{}{dialect}}, where \texttt{dialect} specifies one of the four target dialects. We also employed the corresponding Arabic prompt for generation:

\begin{RLtext} \texttt{ اكتب تغريدة بال\LR{\{dialect\}}}\end{RLtext}

\paragraph{EASC Articles Summaries}

We used the Essex Arabic Summaries Corpus, a dataset featuring 153 Arabic articles sourced from Wikipedia, Alrai Newspaper, and Alwatan Newspaper. Each article is accompanied by five human-written summaries, from which we sampled one per article. The corpus spans a diverse range of topics, including art, education, politics, health, science, and finance. To generate machine-produced counterparts, we used \gptfouro with the prompt below. The prompt emphasizes that the generated summary should be highly informative, preserving all key ideas while remaining concise, coherent, and to the point, without introducing unnecessary details or redundancy.

\begin{quote}
    \small
    \begin{RLtext}
            \texttt{قم بتلخيص هذا المقال محافظا على اهم النقاط ملتزما الايجاز و الدقة\LR{\{article\}:}}.
    \end{RLtext}

\end{quote}

\paragraph{Youm7 News Articles}
The Kalimat dataset consists of articles sourced from the Egyptian news journal \textit{Youm7}. This dataset comprises a total of 21,000 articles across various topics, providing a diverse base for text generation and analysis. For our study, we sampled 1,000 news and generated using GPT-4o (a multilingual LLM) and AceGPT (a Arabic-centric LLM). 

The text generation process was guided by a carefully crafted prompt to simulate the typical writing style of \textit{Youm7} articles. This prompt instructed the models to act as a professional journalist, ensuring coherence and alignment with the original content’s structure and tone. Below is the prompt used for this task:
\begin{quote}
    \small
    \begin{RLtext}
        \texttt{تصرف كأنك كاتب أخبار محترف، واكتب مقالًا باللغة العربية يتألف من حوالي \LR{\{word\_count\_rounded\}} كلمة، بعنوان \LR{\{title\}}. اعتمد في أسلوبك على طريقة كتابة مؤلفي \LR{\{source\}}، مع الأخذ في الاعتبار أن الموضوع يندرج تحت فئة \LR{\{topic\_arabic\}}.}
    \end{RLtext}
\end{quote}




The goal of the generated samples was to maintain narrative consistency and topical relevance while adhering to the typical writing style and article structure commonly found in \textit{Youm7}.

\paragraph{SANAD News}
\label{sanad_before_prompt_eng}
The SANAD news dataset \cite{EINEA2019104076} is a large-scale collection of Arabic news articles written in MSA, designed to support a variety of NLP tasks. It comprises over $190{,}000$ articles from three major news outlets: AlKhaleej, AlArabiya, and Akhbarona, and is released under the CC-BY-4.0 license.

We generated a total of $500$ news articles using titles from the SANAD dataset, sampling $100$ articles for each of five categories: finance, politics, sports, medicine, and technology. The articles were generated using OpenAI’s GPT-4o, and the prompt used for generation is shown below:

\begin{quote}
    \small
    \begin{RLtext}
    \texttt{تصرف كأنك كاتب أخبار محترف ، و اكتب مقالًا باللغة العربية يتألف من حوالي \LR{\{word\_count\_rounded\}} كلمة، بعنوان \LR{\{title\}}. اعتمد في أسلوبك على طريقة كتابة مؤلفي \LR{\{source\}}، مع الأخذ في الاعتبار أن الموضوع يندرج بشكل عام تحت فئة \LR{\{topic\_arabic\}}.}
\end{RLtext}
\end{quote}

In this prompt, we instruct the LLM to act as a professional news writer and generate an Arabic news article of approximately \texttt{word\_count\_rounded} words based on the title \texttt{title}. The model is guided to follow the writing style of authors from \texttt{source}, while ensuring the content aligns with the broader topic category \texttt{topic\_arabic}. Here, \texttt{word\_count\_rounded} denotes the word count of the corresponding human-written article, rounded to the nearest hundred. \texttt{source} refers to the publisher associated with the title in the SANAD dataset, and \texttt{topic\_arabic} corresponds to one of the five predefined categories, selected to match the topic of the article.

\subsection{Chinese}
We collected three types of Chinese text: Zhihu question answering, high school student essays, and government reports.

\paragraph{Zhihu QA}
We used a Chinese question-answering dataset from Zhihu, a popular platform where users post, answer, like, and share questions. From this dataset, we randomly sampled 588 examples.\footnote{\url{huggingface.co/datasets/zirui3/zhihu_qa}}
 These cover diverse topics such as life, technology, art, science, emotion, law, fashion, humor, and China-specific culture.
For emotion-rich scenarios, we additionally curated questions across ten themes, including breakups, quarrels, happiness, marriage, divorce, depression, inferiority, parents, forgiveness, and future concerns (10 questions per theme), emphasizing emotionally expressive and occasionally humorous responses.
We then collected corresponding answers generated by a multilingual LLM (\gptfouro) and a Chinese SOTA model (\qwenturbo).

\paragraph{Junior and Senior High School Student Essays}
We sampled 600 Chinese student essays from \citet{song-etal-2020-multi}, ensuring a balanced distribution across essay genres and student grade levels.\footnote{Each instance includes the essay, its genre (e.g., narrative, argumentative), a quality rating (bad to excellent), and the student grade (7–9 for junior, 10–12 for senior high school).}
Since the dataset does not include corresponding prompts or essay topics, generating LLM counterparts is non-trivial. To address this, we collected: (i) 376 essay prompts from Gaokao exams (1977–2024) and junior high school assignments, and (ii) 51 (prompt, essay) pairs with high-quality human-written essays.\footnote{\url{https://docs.google.com/spreadsheets/d/1iwtnQkamxoqnThWUT9F007WUHm1XmNESAwSVFM8xNdU/edit?usp=sharing}}

We sampled 300 prompts from (i), prioritizing recent years, combined them with all examples from (ii), and generated machine-written essays using \claude-3.5-Sonnet and \chatglmfour to ensure diversity and quality.

\paragraph{Government Reports}
The GovReport dataset, derived from the MNBVC GovReport subset~\cite{mnbvc}, contains titles and report bodies from various institutions, including schools, corporations, and local government entities, without additional annotations. We randomly sampled 500 entries to construct a diverse evaluation set.
To encourage output diversity, we designed eight prompts, including continuation tasks (based on titles or opening sentences) and rephrasing tasks (based on full reports). These prompts, written in Chinese, were alternated to guide LLM generation. We collected outputs from GPT-4o, Baichuan2-13B-Chat, and ChatGLM3-6B.

\subsection{English Peer Meta Review}
We randomly sample 600 reviews (NeurIPS 2021-2022) from PeerSum dataset~\citep{peersum_2023}. 
Peersum crawled both meta reviews and reviews from each reviewer and rebuttal from authors from openreview\footnote{\url{https://openreview.net/}}. 
We leveraged this dataset to generate meta reviews based on comments from other reviewers and compared with human-written meta reviews.
We generate meta reviews using \texttt{\gptfouro-2024-08-06} and \texttt{claude-3.5-sonnet-20240620} respectively, with prompt ``Generate a meta review based on the reviews' opinions and authors' rebuttal to make the final decision on whether the paper should be accepted: \texttt{\{Reviews\}}$\backslash $n$\backslash$n Meta review:'', where ``\texttt{Reviews}'' contains both the reviewers' review and authors rebuttals. 
We sampled 400 machine-generated meta reviews, paired with humans, and asked annotators to discern.

\subsection{Hindi News}
The BBC Hindi news article dataset comprises a diverse selection of around 4,000 Hindi news articles sourced from the BBC Hindi website, covering a wide range of topics and categories. Each article includes three primary components: headline, main content and the thematic category of the article.

We sampled 600 articles from the original dataset, focusing on articles that span multiple prominent topics in Indian news to ensure a balanced representation. These articles were selected randomly to avoid any thematic bias and provide a broad scope for comparison.
We generated 600 machine-written samples using GPT-4o model, employing a prompt that guided the model to write concise and formal articles: ``Here is a news headline: '{headline}' and the content: '{content}'. Write a machine-generated version of the news based on this headline''.


\subsection{Italian DICE News}
For Italian, we used DICE~\citep{bonisoli2023} --- local news from La Gazzetta di Modena ($\approx 10,000$ samples) licensed as CC-by-nc-sa.
We sampled 300 news, where the original text has at least 1000 characters. 
We applied three large language models: Llama-3.1-405b-instruct, \gptfouro, and Anita (an Italian fine-tuned Llama-3.1-8B: \citet{polignano2024advanced}). 


When generating in Italian, we use one of 4 possible system prompts as \tabref{tab:Italian-prompt}.
The model is always prompted as a journalist but the mother tongue and the newspaper scope may vary lightly.
The general format is the following:
\begin{quote}
\textbf{System:} You are an Italian journalist writing for a national newspaper focusing on criminal events
happening in the area surrounding Modena. \\
\textbf{User:} Write a piece of news in Italian, that will appear in a local Italian newspaper and that has the following title: ...
\end{quote}

\begin{table*}[ht]
\centering
\begin{tabular}{|p{7cm}|p{7cm}|}
\hline
\multicolumn{2}{|c|}{English-first LLMs} \\
\hline
\multicolumn{2}{|c|}{System Prompt} \\
\hline
    You are an Italian journalist writing for a national newspaper focusing on criminal events happening in the area surrounding Modena & You are an Italian journalist writing for a local newspaper focusing on criminal events happening in the area surrounding Modena \\
    \cline{1-2}
    You are an Italian-French journalist writing in Italian about criminal events happening in the area surrounding Modena & You are an Italian-American journalist writing for a local newspaper focusing on criminal events happening in the area surrounding Modena \\
\hline
\multicolumn{2}{|c|}{User Prompt} \\
\hline
    Write a piece of news in Italian, that will appear in a local Italian newspaper and that has the following title: & Write a piece of news in Italian, that will appear in a local Italian newspaper and that has the following title: \\
    \cline{1-2}
    Write a piece of news in Italian, that will appear in a national Italian newspaper and that has the following title: &     Write a piece of news in Italian, that will appear in a local Italian newspaper and that has the following title: \\
    \hline
\multicolumn{2}{|c|}{Italian-first LLMs} \\
\hline
\multicolumn{2}{|c|}{System Prompts} \\
    \cline{1-2}
    Sei un giornalista italiano che che scrive per un giornale nazionale focalizzandosi su eventi criminali che accadono a Modena & Sei un giornalista italo-francese che scrive in italiano su eventi criminali che accadono a Modena \\
    \cline{1-2}
    Sei un giornalista italiano che scrive per un giornale locale focalizzandosi su eventi criminali che accadono a Modena & Sei un giornalista italo-americano che scrive per un giornale locale focalizzandosi su eventi criminali che accadono a Modena \\
    \cline{1-2}
\multicolumn{2}{|c|}{User Prompts} \\
    \cline{1-2}
    Scrivi un articolo di giornale in italiano. L'articolo sarà pubblicato su un giornale locale e avrà il seguente titolo: & Scrivi un articolo di giornale in italiano. L'articolo sarà pubblicato su un giornale nazionale e avrà il seguente titolo: \\
    \cline{1-2}
    Scrivi un articolo di giornale in italiano. L'articolo sarà pubblicato su un giornale locale e avrà il seguente titolo: & Scrivi un articolo di giornale in italiano. L'articolo sarà pubblicato su un giornale locale e avrà il seguente titolo:\\
\hline
\end{tabular}
\caption{Italian machine generation prompts.}
\label{tab:Italian-prompt}
\end{table*}

\subsection{Japanese News}
We randomly sample 300 news articles of the BBC news from the XLSUM dataset \citep{hasan-etal-2021-xl}.
XLSUM has title, content, and summary of a news article.
We generate the corresponding news content using \texttt{\gptfouro-2024-08-06} based on the article titles.
For both text generated with \gptfouro and the human-written text, we remove obvious formatting indicators (e.g., line breaks and template messages at the beginning and end of the texts).

We generate 75 articles for each of the four settings: simple prompt zero-shot, diverse expression zero-shot, content-rich zero-shot, and few-shot.
Simple prompt zero-shot, diverse expression zero-shot, and content-rich zero-shot use the following instructions, respectively:

``\CJK{UTF8}{min}{次のニュースタイトルに合わせたニュース記事を生成してください。}'' (Please generate a news article that matches the following news title.), ``\CJK{UTF8}{min}{次のニュースタイトルに合わせたニュース記事を多様な表現を使用して生成してください。}'' (Please generate a news article that matches the following news title, using diverse expressions.), and ``\CJK{UTF8}{min}{次のニュースタイトルに合わせたニュース記事を生成してください。このとき生成するニュース記事には、誰かに対するインタビューや実際の出来事を組み込んでください。}'' (Please generate a news article that matches the following news title. When creating the article, include interviews with individuals or actual events.).
The few-shot uses the same instruction as the simple prompt zero-shot.
Add ``\CJK{UTF8}{min}{ニュースタイトル:}'' (news title:) and ``\CJK{UTF8}{min}{ニュース記事:}'' (news article:) at the beginning of the news title and the content, respectively.
We sampled three contents as examples for few-shot, and use the same examples for all generations.

\subsection{Kazakh Wikipedia}
KazQad is a closed question-answering dataset focused on the Kazakh~\citep{kazqad}. It contains 5,000 distinct passages covering 1,700 topics derived from Kazakh Wikipedia. These passages span a variety of domains, including art, science, history, sports, and other general topics.
 
We randomly selected 300 titles along with their corresponding paragraphs, ensuring that each paragraph contained at least 3-5 sentences. Additionally, the sampled texts were cleaned and merged by title to increase the length of the texts. This step was necessary because some samples in the original dataset contained extraneous elements such as references, markdown formatting symbols, and other unnecessary characters.
For the machine-generated data, we used the GPT-4o model to generate passages based on the sampled titles. The generation process was initiated with the following prompt: ``Please, write one paragraph about the following topic in Kazakh: [title].''

\subsection{Russian}
We generated machine text for both news and summaries for Russian.

\paragraph{News}
Based on Lenta.ru from Corus with 800K news samples, we sampled 50 cases per topic from six topics: Russia, World, Economy, Sport, Culture, and Science \& Technology. 

We prompted GPT-4o and Vikhr-Nemo-12B-Instruct-R-21-09-24 to generate text: \foreignlanguage{russian}{''Напиши новость в области "{topic}" с сайта lenta.ru используя заголовок {title}. Ты должен генерировать новость без заголовка. Новость: ''}

\paragraph{Academic Article Summaries} 
Based on ai-forever/ru-scibench-grnti-clustering-p2p with 31K samples, we sampled 30 summaries per topic from these topics: 'Psychology', 'Mechanical Engineering', 'Agriculture and Forestry', 'Geology', 'Biology', and 'Energy', and generated machine counterparts using GPT-4o and Vikhr-Nemo-12B-Instruct-R-21-09-24.  
We used the following prompt: \foreignlanguage{russian}{''Напиши краткое содержание для статьи в области "{topic}" используя заголовок {title}. Ты должен генерировать краткое содержание без заголовка. Содержание:''}

\subsection{Vietnamese}

We generated data from news and Wikipedia.

\selectlanguage{Vietnamese}
\paragraph{Vietnamese News} 
The dataset is crawled from Lao Động newspaper\footnote{\url{https://www.kaggle.com/datasets/phamtheds/news-dataset-vietnameses}} before May 2022 (before the releases of all LLM that support Vietnamese). The dataset contains 290,282 articles with a headlines and a summary of an article in various topics, such as politics, lifestyles, legal, etc. We cleaned the data by removing some missing values and too short summaries (less than 20 words), then randomly selected 600 examples to generate the corresponding 600 machine-generated summaries with GPT-4o using title of articles. The prompt for generation process was this: 
\begin{quote}
    \textit{Bạn là một nhà báo Việt Nam chuyên viết những mẩu tin tóm gọn cho các bài báo bằng cách sử dụng tiêu đề của chúng. Hãy viết cho tôi một đoạn tóm tắt bài báo tiêu đề dưới đây:}
\end{quote}
which means: \textit{You are a Vietnamese journalist who writes summaries for articles using their headlines. Please write me a summary of the article with the following headline:}

\paragraph{Vietnamese Wikipedia} 
We randomly crawled 600 sites from Vietnamese Wikipedia\footnote{\url{https://vi.wikipedia.org/}} with its ID, title, and the introduction part of each topic. For our studies, we generated 600 introduction given a subject using GPT-4o with the prompt:
\begin{quote}
    \textit{Bạn là một nhà đóng góp cho Wikipedia tiếng Việt. Hãy viết cho tôi một đoạn giới thiệu ngắn gọn bằng tiếng Việt về chủ thể bên dưới để đăng trên trang Wikipedia. Lưu ý bắt buộc viết với khoảng \texttt{word\_count} từ.}
\end{quote}
which means \textit{You are a contributor to Vietnamese Wikipedia. Please write me a brief introduction in Vietnamese about the subject below to post on the Wikipedia page. Note that it must be written within \texttt{word\_count} words:}

In this prompt, to avoid the newly generated text has too long passage, since GPT-4o tends to write much longer than human does for a Wikipedia topic, we set the word limits \texttt{word\_count} as word count of the original one.
\selectlanguage{English}



\section{Distinguishable Signals}
\label{sec:distinctionfactor}
This section elaborates detection setting, accuracy and distinction signals summarized by 18 annotators for 16 datasets one by one.


\subsection{Arabic}
\label{sec: arabic_insights}
\paragraph{Arabic Dialect Tweets}
The annotation was conducted under the setting III. Single-binary, where a total of 900 tweets across four dialects were analyzed, with ratio of machine-generated vs. human-written text as 2:1 (GPT-4o and Qwen2-7.5B).

The annotators were not very confident about their choices when deciding whether the data is human-written or not. They achieved an overall accuracy of 50.06\%, highlighting the difficulty of distinguishing machine-generated content in short texts. 

Overall, 64.17\% of human-written tweets were correctly annotated, 36\% human-written tweets were mis-identified as machine-generated.
\tabref{tab:arabic-dialect_tweet-accuracy} presents the human detection accuracy on human text and machine-generated tweets across four dialects, machine-generated text is harder than human text to discern. \gptfouro outputs are more similar to human text, thus more difficult to distinguish compared to \qwentwo.

Since LLMs effectively replicate native speaker vocabulary, human detection cannot rely on distinguishing lexical distinctions. A key indicator of machine-generated text was the use of emojis in an overly formulaic manner. Additionally, some tweets addressed topics that would typically be considered trivial or unlikely for humans to post, further suggesting machine generation.  
Machine-generated tweets also exhibited unnatural tones, misused native expressions, or contained incomplete content. A notable issue in \qwentwo's output was the inclusion of words from other languages within sentences and the mixing of dialects, which is uncommon in natural usage.



\paragraph{EASC Articles Summaries}

We sampled 100 (human, \gptfouro) pairs to identify which text is written by human, achieving 82\% accuracy.
The annotator differentiated human-written text from the machine-generated text based on five empirical distinguishable signals.
\begin{itemize}
    \item \textbf{Informative:} LLM summaries are more informative, while humans may overlook key points due to emotional biases, personal perspectives, or incomplete understanding of articles.
    \item \textbf{Abstract and concise:} LLMs present ideas in a more abstract and concise way, without emphasis on specific points. Human summaries often inject personal opinions and beliefs.
    \item \textbf{Religious language:} Humans can accurately use religious language, whereas LLMs generate text in standard language.
    \item \textbf{Prompt reflection:} LLMs often start with words from the prompt, such as
    \begin{quote}
    \small
    \begin{RLtext}
            \texttt{هذا المقال}
    \end{RLtext}
    \end{quote}
    \item \textbf{Formatting hints} Human-written summaries contain typos or grammar errors. Machine-generated text includes markdown elements, making it more easily detectable as machine-generated.
\end{itemize}


\paragraph{Youm7 News Articles}

We created 1,000 (human, \gptfouro) paired examples to identify which text is written by humans, achieving 92.7\% accuracy. A key finding was that human-written paragraphs were typically presented as a single continuous block, whereas GPT-generated articles were segmented into smaller sections. This structural difference served as a key cue for identifying machine-generated content. No significant variations were observed in thematic patterns or narrative sequences, but paragraph segmentation emerged as a key indicator, highlighting the role of structural features in detection.

\paragraph{SANAD News}
Using the SANAD Arabic news dataset in MSA, we evaluated 100 samples to identify human vs. GPT-4o text, achieving $100\%$ annotation accuracy under setting I. Pair-binary. We identified several key features that differentiate machine-generated text from human-written text.
\begin{itemize}
    \item \textbf{Markdown presence:} 
    Machine-generated text often contains markdown, which is absent in human-written text.

    \item \textbf{Formatting style:} 
    Machine-generated text is consistently formatted into structured paragraphs, whereas human text tends to appear as large, unstructured blocks of text.

    \item \textbf{Content density:} 
    Human text is richer in factual information, while machine-generated text includes generic statements. For instance, phrases such as 
    ``Saudi Vision 2030'' are more frequent in machine-generated text. 

    \item \textbf{Source attribution issues:} 
    Machine-generated text often contains source attributions, but these are sometimes incorrect or against the prompt. 

    \item \textbf{Presence of numbers and specific meta data:} 
    Human-written text contains more supporting numerical details, including URLs, phone numbers, currency exchange rates, dates, and amounts, which are not as frequent in machine-generated text.

    \item \textbf{Use of English terms:} 
    Human text includes sporadic English terms, especially in specialized contexts, while machine text remains fully in Arabic.

    \item \textbf{Hashtags and social media elements:} 
    Hashtags are commonly found in human-written text, whereas machine-generated text lacks such social media elements.

    \item \textbf{Narrative structure:} 
    Human-written text follows a narrative structure with a sequence of events, dates, and timelines. On the other hand, machine-generated text tends to resemble an essay on a given topic rather than a chronological news report.

    \item \textbf{Formatting consistency:} Human text formatting varies significantly, often appearing as inconsistent or messy blocks of text with irregular spacing or newlines. We attribute this apparent inconsistency to the fact that the human texts are composed by a large number of authors, while machine text is written by a single model which provides text that is more polished and consistent.

    \item \textbf{Readability and grammar:} 
    Machine-generated text is generally more readable and grammatically correct, while human text may contain stylistic inconsistencies.
\end{itemize}

\subsection{Chinese}
\label{sec:chinese-distinguishable-signal}
\paragraph{Zhihu QA}
The detection involves six unique individuals, with three female and three male annotators. 
We pair (human, \gptfouro) and (human, \qwenturbo), and annotate both under setting I, achieving the average accuracy of 1.0 and 0.98 respectively. The distinguishable factors are summarized below.
\begin{itemize}
    \item \textbf{Humans share personal experience and feelings.} For emotion-rich questions, humans provide empathic comforts by sharing their real personal experiences. However, narratives or stories generated by models tend to appear contrived and artificial, making it easy for humans to detect that they are not grounded by facts, similar to stories created by kids. Model responses thus typically offer solutions, but they are theoretically sound, while often lack practical applicability.    
    
    \item \textbf{Human answers can be informal, mean and sharp.} 
    Human responses sometimes are mean, strongly opinionated, and influenced by personal biases, reflecting a more self-centered perspective. In contrast, LLMs provide general and less engaging information, but often attempting to help users and offering problem-solving assistance. Some individuals appreciate human answers for their authenticity and directness, others may find them offensive.
    
    \item \textbf{Different Intent.} Zhihu users prioritize expressing their feelings and opinions, rather than assisting and directly addressing the seeker's needs. So they often do not respond to the question directly. Though LLMs aim to assist but typically offer general information related to the entities or concepts in the question, rarely responding with direct and sharp answers.
   
    \item \textbf{Human answers can be extremely short or long.} Some human answers are excessively long and lack proper paragraph segmentation, making them harder to read. LLM-generated responses are generally well-segmented and structured with bullet points or listed points with bold subtitles.

    \item \textbf{LLM responses lack deviation.} LLMs adhere rigidly to instructions, presenting limited flexibility in their responses. When prompted to answer emotion-rich questions with greater empathy, their generative patterns remain predictable, often relying on superficial expressions such as more emoji inserted instead of deeply integrating empathy into the content. For instance, responses like \textit{I'm sorry this happened to you; you should probably consult a professional}, offer minimal support and feel unhelpful.

    \item \textbf{Other indicators} Human answers can contain timeline of answer update and references. 
\end{itemize}

\textit{Improved Prompts}
We carefully designed prompts using a few different user personas: one with a positive attitude, another with a rational and realistic outlook, and a humorous one. For emotion-rich questions, we also applied \gptfouro utilizing \ecot, a framework designed to enhance LLMs’ emotional and empathetic responses~\cite{li2024ecot}. However, the responses are often brief, with emojis but minimal information, which limits their usefulness.

\paragraph{High School Student Essays}
We perform human detection for student essays in two settings.
One NLP postdoc who is a Chinese native speaker did the detection under the setting I for 102 parallel pairs ($hwt$, $mgt_1$) and ($hwt$, $mgt_2$), obtaining the accuracy of 98\%.
For 600 non-parallel cases under the setting II. Pair-four-class, with 150 pairs for each class, another NLP postdoc and two NLP PhD students participated in annotations. They achieved an accuracy of 0.96, 0.96 and 0.99 respectively. 

During the annotation of high school student essays, we identified four signals that make machine-generated texts easily recognizable:
\begin{itemize}
    \item \textbf{Title:} From the perspective of formatting, machine text tends to begin by ``title: \cn{《xxx》} or title: xxx'', and have newlines between paragraphs while humans does not have.
    
    \item \textbf{Formulaic structure:} From the structure and the content of essays, the structure of machine-generated essays is generally formulaic. They often adopt an argumentative style that begins paragraphs with phrases such as ''first, then, moreover, additionally, finally, and overall'' (i.e., \cn{首先，其次，然而，总之，最后}), while HWT is more flexible and the styles are more diverse. Some MGT even use bullet points in an essay, which is rare in human text.

    \item \textbf{Sentiment:} Machine-generated essays tend to adopt a more neutral or positive tone, generally expressing less emotion compared to human-written essays. While real students often express a range of emotions, including sadness, anger, confusion, and a sense of being lost. These emotions reflect the authentic feelings of young people at that age.
    
    \item \textbf{Style:} Machine-generated content may incorporate elements from other genres, such as official document styles, or may unexpectedly switch to multilingual content, e.g.,~outputting an English paragraph during Chinese content generation.
\end{itemize}

\textit{Original vs. Improved Prompt}
Based on the findings above, we further refined the prompts in the following ways: (1) instructing the model to avoid outputting titles, as these often serve as clear detection cues; (2) discouraging excessive use of connecting words like 'first of all,' 'secondly,' 'then,' and 'finally'; and (3) preventing the mixing of languages other than Chinese.

\paragraph{Government Report}
Under setting IV. Triplet-three-class, we presented human-written texts vs. two model outputs across 500 samples to a native Chinese-speaker (postdoctoral researcher specializing in NLG), and asked the annotator to identify which text is human-written. 
The annotator achieved an accuracy of 97.2\%. Human-authored texts were typically longer and contained richer details, often covering multiple topics, while machine-generated texts were generally shorter, lacking noticeable rhythm variations. 

Additionally, machine-generated texts occasionally included distinct symbols, such as bold formatting and English words.
Errors in distinguishing humans from machine-generated texts primarily occurred when the machine-generated outputs were of similar length to human-written ones.

\subsection{English Peer Meta Review}

Human detection accuracy is 99.75\%.
Before seeing labeled samples, the annotator found the distinction was obvious. After reviewing a few examples, the annotator was extremely confident in distinguishing human-written from machine-generated text based on indicative features of MGT. 

Machine-generated peer reviews exhibited distinct, predictable patterns. They frequently followed a structured format, often providing explicit decisions as ``Decision: Accepted'', which was rare in human-written reviews. MGT also commonly used headings such as ``Strengths'' and ``Weaknesses''. MGT tended to explicitly summarize multiple reviews while the human reviewers rarely did. An example of MGT is like ``Based on the reviews, this paper presents methods for embedding numerical features to improve deep learning models for tabular data. The key points are:''.

MGT additionally presents a notable preference for bullet points, even when a paragraph format was feasible, with minimal variation in bullet style. MGT responses tend to be more uniform in length and generally longer than human reviews. In contrast, human-written meta reviews lacked a fixed structure, rarely used bullet points, and exhibited greater variation in length, though they were typically shorter than MGT.

\subsection{Hindi News}
An overall accuracy of 85.17\% was achieved in distinguishing between the machine-generated and human-written Hindi BBC news articles.
The analysis reveals key stylistic and content-based distinctions.
One significant observation was that machine-generated news content tended to be more concise, presenting less overall information compared to its human-written counterpart. Additionally, machine-generated articles included fewer mentions of names of persons and specific dates, elements that are typically embedded within human-authored news to enhance credibility and specificity.
Furthermore, stylistic disparities were evident, particularly in the absence of colloquial language elements commonly used by human authors. 

Human-written articles often incorporate idioms, culturally significant phrases, and even a blend of Urdu vocabulary, reflecting the linguistic diversity and nuance of the Hindi language. These elements, however, were noticeably missing in machine-generated content, which instead adhered strictly to the main topic, with minimal linguistic or thematic deviations. This adherence to topic, while adding to clarity, lacked the depth and regional authenticity often present in human-created Hindi news content. Also there was use of English text in the machine generated version.

\subsection{Italian News}
Based on human text from DICE dataset~\citep{bonisoli2023}, we randomly selected machine-generated text from GPT-4o, Anita \citep{polignano2024advanced} and Llama3-405B \citep{llama3} outputs, resulting in total of 300 (human, MGT) pairs.
A native Italian speaker was asked to choose which text is human-written. Detection accuracy is 88.0\% for Anita, 99\% on Llama3-405B and 100\% for GPT-4o.


We identified the following distinguishable signals. 
When generating news, all models especially the larger ones, Llama-405b and GPT-4o have consistent formatting, e.g., article title is always enclosed in markdown-like double asterisks ``**'', the city where the issue happens is always mentioned first. These cues are immediately recognizable. After seeing only 2-3 examples, human annotator was confident to make decisions and obtain close to 100\% detection accuracy.


\subsection{Japanese News}
\label{japanese_news_distinctive_clues}
We divided 300 pairs into two groups and two annotators independently annotated them, to identify which text was a human-written BBC news article.
Human annotators achieved an average accuracy of 62\%. 
We identified several distinctive characteristics of LLM-generated texts:
\begin{itemize}
    \item \textbf{Style:} Some texts failed to match news writing styles.
    For instance, in Japanese, writing typically either uses the \textit{desu/masu} or \textit{dearu} style.
    While news articles conventionally use the \textit{dearu} style, LLM-generated texts often use the \textit{desu/masu} style.
    \item \textbf{High reliance on headlines:} The opening sentences frequently relied on the title (headline), either through direct repetition or close paraphrasing.
    \item \textbf{Formulaic phrase:} LLM outputs sometimes contained formulaic phrases like \textit{This article is provided as fiction} or \textit{Here are five reasons: 1. ...}.
    \item \textbf{Typos and grammar issues:} We observed various typographical errors and fluency issues, such as \textit{This four has shocked the entire UK} or \textit{An event that shook the air in Myanmar}.
\end{itemize}
These findings suggest that LLMs have not yet fully mastered the stylistic conventions of news writing.
Explicit instructions that indicate the domain-specific format and style may help the LLM output.

\subsection{Kazakh Wikipedia}
A native Kazakh speaker and a person proficient in Kazakh annotated 300 (human, \gptfouro) pairs under the setting I, with detection accuracy of 79.67\%.
We summarized the following features of MGT.
\begin{itemize}
    \item The generated text lacks diversity in expression, often using repetitive sentence patterns and predictable phrasing, such as frequently ending with ``bolyp tabylady'', which contributes to a mechanical and formulaic feel.
    \item The generated text rarely includes concrete facts, such as numbers or years, and when they do appear, their occurrence is minimal.
    \item The text also frequently includes flattering language, which can be another signal of its artificial nature.
    \item Human-written texts include more Kazakh-specific cultural references where applicable, making them more relatable and authentic, which serves as a clear signal of human-generated content over LLM-generated text.
\end{itemize}

\subsection{Russian}
\paragraph{News Articles}
A native Russian speaker (NLP PhD student) annotated 300 examples under the setting I, where machine-generated text was randomly selected from either GPT-4 or Vikhr outputs. Accuracy is 100\%. 
We found that human texts often include specific details like exact dates, numbers, names of people, places, or things (especially those not mentioned in the title or from ordinary backgrounds), as well as ages and other specifics. Additionally, human-written texts tend to reference sources, which is a strong indicator.


\paragraph{Academic Article Summaries}
A native Russian speaker (NLP PhD student) was asked to identify whether the given text is human-written or machine-generated under the setting III. Single-binary. A text is randomly chosen from human-written, GPT-4, or Vikhr outputs (300 samples).
The overall accuracy is 80\% (Psychology: 80.0\%, Mechanical Engineering: 74.0\%, Agriculture and Forestry: 74.0\%, Geology: 86.0\%, Biology: 86.0\%, and Energy: 80.0\%).
There are several indicators. Machine-generated texts, especially from Vikhr, often begin with a paraphrased version of the title, which is a strong indicator. Human-written texts typically contain details such as numbers, references, and names. In some Vikhr outputs, sentences may lack coherence, and even after preprocessing, certain artifacts remain visible in the text.


\subsection{Vietnamese}
We annotated 600 news summaries and 600 Wikipedia introduction passages in the setting of I. Pair-binary. 
We obtained accuracy of 80.33\% on news summaries, but random guess 50.67\% on Wikipedia text, showing minimal differences between human-written and machine-generated Wikipedia passages.

A key finding was that, for news articles, human tends to provide more details to support the statement (such as date, location, and other specific information), while LLMs tend to offer a general summary. Besides, GPT-4o usually uses short sentences in summaries, but Vietnamese journalists usually write long sentences.


For Wikipedia passages, it is much harder for humans to distinguish since GPT-4o was well-trained on Wikipedia data. The differences between human-written and machine-generated Wikipedia text are subtle, though some distinctions remain observable in tone, emphasis, and content selection. For example, when GPT-4o was asked to write about Hai Phong city, city demographics and geography are expected, yet it generated a review-like sentence:
\textit{Modern infrastructure, along with open investment policies, have helped Hai Phong become an attractive destination for domestic and foreign investors, contributing to the city's sustainable development.}
This resembles promotional or evaluative writing rather than the neutral, descriptive style typical of Wikipedia introductions.



\section{Fill in the Gap by Prompting?}
\label{sec:fill-gap-prompts}

\begin{table*}[t!]
    \centering
    \small
    \begin{tabular}{llr cll c}
    \toprule
    \textbf{Language} & \textbf{Source/Model} & \textbf{\#Example} & \textbf{\#Annotator} & \textbf{Yes} & \textbf{Partially} & \textbf{No}  \\
    \midrule
    \multirow{4}{*}{Arabic} 
    & Dialect Tweet & 200 & 1 & 106 & 58 & 36 \\
    & ESAC Summary & 130 & 1 & \multicolumn{3}{c}{82.0\% $\rightarrow$ 62.7\%} \\
    & Youm7 News & 200 & 1 & \multicolumn{3}{c}{92.7\% $\rightarrow$ 52.0\%} \\
    & SANAD News & 200 & 1 & \multicolumn{3}{c}{100\% $\rightarrow$ 66.0\%} \\
    \midrule
    \multirow{6}{*}{Chinese} 
    & Zhihu-QA & 100 & 3 & \multicolumn{3}{c}{99.7\% $\rightarrow$ 94.3\%} \\
    & Zhihu-QA & 200 & 1 & 32 & 70 & 98  \\
    & \multirow{3}{*}{Student essay}  & \multirow{3}{*}{200} & \multirow{3}{*}{3} 
    &  86 & 36 & 78 \\
    & & & & 84 & 36 & 80 \\
    & & & & 45 & 23 & 132 \\
    & Government Report & 200 & 1 & 59 & 38 & 103 \\
    \midrule
    English & Peersum & 200 & 1 & 2 & 113 & 85   \\
    \midrule
    Hindi & News & 200 & 1 & \multicolumn{3}{c}{85.2\% $\rightarrow$ 66\%}  \\
    \midrule
    \multirow{3}{*}{Italian} 
    & DICE News (Anita) & 300 & 1 & \multicolumn{3}{c}{88\% $\rightarrow$ 81\% } \\
    & DICE News (Llama3-405B) & 300 & 1 & \multicolumn{3}{c}{99.7\% $\rightarrow$ 84\% } \\
    & DICE News (GPT-4o) & 300 & 1 & \multicolumn{3}{c}{100\% $\rightarrow$ 85\%}  \\
    \midrule
    Japanese & News & 200 & 2 & \multicolumn{3}{c}{ 86.4\% $\rightarrow$86\% } \\
    \midrule
    Kazakh & Wikipedia & 200 & 2 & 105 & 89 & 6 \\
    \midrule
    \multirow{2}{*}{Russian} 
    & News & 200 & 1 &  \multicolumn{3}{c}{100\% $\rightarrow$ 86.5\%} \\
    & Academic summary & 200 & 1 & \multicolumn{3}{c}{80\% $\rightarrow$ 69\%} \\
    \midrule
    \multirow{2}{*}{Vietnamese} 
    & Wikipedia & 600 & 1 & \multicolumn{3}{c}{ 50.7\% $\rightarrow$ 47.3\% } \\
    & News & 600 & 1 & \multicolumn{3}{c}{ 80.3\% $\rightarrow$ 63.2\% } \\
    \midrule
    Total & -- & 4,730 & 25 & \multicolumn{3}{c}{ \textbf{87.6\%} $\rightarrow$ \textbf{72.5\%} }  \\ 
    \bottomrule
    \end{tabular}
    \caption{Human detection accuracy differences on original vs. improved generations, and survey distribution evaluating whether the new generations fill the gap: Yes, Partially or No.}
    \label{tab:fill-gap-survey}
\end{table*}

In this section, for each dataset, we first present how we designed the improved prompts, and then elaborate whether the new generations are improved and fill the previously-observed gaps between humans.
Original and improved prompts for all datasets and languages are summarized in \tabref{tab:ori-improved-prompts}, and the detection accuracy on new content and fill-gap survey results are demonstrated in \tabref{tab:fill-gap-survey}.

\subsection{Arabic}
\paragraph{Tweets}

In order to enhance the quality of the machine-generated Arabic tweets, the prompts were refined to better capture human emotions and to generate tweets that can reflect authentic human experiences more naturally and contextually across diverse situations. One such improved prompt was: ``\emph{Generate a random tweet in Arabic. Use {dialect} dialect, use human emotions and experiences. Output the tweet only}'' and its corresponding Arabic translation is as follows:
\begin{quote}
    \small
    \begin{RLtext}
    اكتب تغريدة عشوائية باللغة العربية. استخدم اللهجة \LR{\{dialect\}}وعبّر عن مشاعر وتجارب إنسانية
    \end{RLtext}
\end{quote}

These adjustments addressed critical gaps appeared in prior outputs. The newly-generated tweets touched on relatable human topics and expressed genuine emotions tied to daily experiences. We used \gptfouro for generation as it yielded promising results, while Qwen is underperformed. As mentioned in \appref{sec: arabic_insights}, Qwen texts often include non-Arabic words within the tweets. 

To evaluate the effectiveness of the improved prompts, 200 sampled tweets were assessed by annotators, determining whether the outputs addressed the identified gaps. The evaluation revealed that 53\% of the tweets fully met the criteria, 29\% partially did, and 18\% did not. However, some limitations were observed: GPT-4o frequently added irrelevant hashtags at the end of tweets, making the outputs identifiable as machine-generated, and the tweets often carried an \textbf{overly optimistic tone}. Even when negative experiences were mentioned, the sentiment leaned toward positivity and new beginnings. 
Overall, while these improved prompts succeeded in filling some previous gaps, new issues emerged, which could be addressed with more precise instructions, such as explicitly avoiding hashtags or adopting a more somber tone when required.

\paragraph{EASC Summary}
Building on the identified discrepancies between human-written and machine-generated summaries discussed in \appref{sec: arabic_insights}, the generation prompt was refined to address these gaps, resulting in the following prompts:
\begin{quote}
    \small
    \begin{RLtext}
            \texttt{قم بتلخيص المقال التالي محاولا محاكاة انسان له اراء و معتقدات فكرية او دينية ملتزما الدقة و الايجاز فمتوسط مستوى البشر\LR{\{article\}:}}.
    \end{RLtext}
\end{quote} 

The data was re-generated using GPT-4o, following the enhanced prompt. This enhancement led to a decrease in annotation accuracy, which dropped to 63\% from 82\%. 
However, there is still typical machine writing style in the new content, like the use of common machine-generated phrases (e.g., \begin{quote}
    \small
    \begin{RLtext}
            \texttt{يتناول هذا المقال...}
    \end{RLtext}
\end{quote} ), clearly indicating that the text was produced by a machine.
Additionally, the newly-generated summaries exhibited several characteristics.
\begin{itemize}
    \item The summaries seem to reflect a particular ideology, often emphasizing religious themes or beliefs, which influenced the tone and content of the generated text.
    \item The generated sentences are not concise, limiting the clarity and precision of the summaries.
    \item The language used to describe events or natural phenomena tendd to be more emotional, with a tendency to exaggerate or reflect personal preferences.
\end{itemize}


\paragraph{SANAD}
Before prompt engineering, the accuracy of annotating which of two texts is machine or human written was 100\%. After prompt engineering, the accuracy dropped to 66\% indicating that prompt engineering works at least partially to bridge the gap between the writing styles of humans and machines. 

We follow the same model choices and data subset outlined in \ref{sanad_before_prompt_eng} but we enhance the prompt as shown below:

\begin{quote}
    \small
    \begin{RLtext}
    تصرف كأنك كاتب أخبار محترف، واكتب مقالًا باللغة العربية يتألف من حوالي \LR{\{word\_count\_rounded\}} كلمة، بعنوان \LR{\{title\}}. احرص على أن يكون المقال غنيًا بالمعلومات والتفاصيل الدقيقة، متضمنًا أرقامًا، تواريخ، ومعلومات كمية إذا كانت مناسبة. تأكد من إدراج روابط، أرقام هواتف، أو أسعار صرف العملات عند الضرورة. استخدم بعض المصطلحات الإنجليزية المتخصصة إن كان السياق يتطلب ذلك، وتجنب الاكتفاء بعبارات عامة مثل 'ويأتي هذا التطور في ظل توجه عالمي...' دون دعم بالأحداث أو التفاصيل. راعِ سرد الأحداث بترتيب زمني واضح يعكس التسلسل المنطقي للوقائع. احرص على أن يكون النص مكتوبًا في فقرة واحدة فقط بغض النظر عن طوله، دون تقسيمه إلى فقرات متعددة. اجعل التنسيق غير مثالي (مثل وجود مسافات غير متساوية) لتعكس طبيعة الكتابة البشرية، ولا مانع من استخدام الهاشتاجات عند الحاجة. يرجى مراعاة أن الموضوع يندرج بشكل عام تحت فئة \LR{\{topic\_arabic\}}. يرجى عدم استخدام \LR{Markdown} نهائيًا.
    \end{RLtext}
\end{quote}

Similar to the original prompt, this prompt instructs the LLM to act as a professional news writer and generate an Arabic news article of approximately `\texttt{word\_count\_rounded}' words with the title `\texttt{title}', ensuring that the article's general topic aligns with `\texttt{topic\_arabic}', one of the five predefined categories.

The enhancements made by this prompt can be summarized as follows:
\begin{itemize}
    \item Produces content rich in information and accurate details.
    \item Includes specific numbers, dates, and quantitative data where applicable.
    \item Encourages the incorporation of specialized English terms when the context requires it.
    \item Advises against vague phrases that are unsupported by concrete events or details.
    \item Requests a clear chronological narration of events.
    \item Requires the article to be written as a single paragraph, regardless of its length.
    \item Allows imperfections in formatting (e.g., uneven spacing) to mimic human-like writing.
    \item Permits the use of hashtags when appropriate.
    \item Prohibits the use of \texttt{Markdown}.
\end{itemize}

The following observations were made from the results of the enhanced prompt:

\begin{itemize}
    \item \textbf{Fake phone numbers} (e.g., 123456789) indicate machine-generated content.
    \item \textbf{Correct phone numbers or URLs} suggest human-generated content.
    \item The use of Arabic \textbf{tatweel} typically indicates human-generated content.
    \item \textbf{Unique names in Arabic} are more likely to be found in human-generated content.
    \item \textbf{Poorly written or formatted texts} are often indicative of human authorship.
    \item \textbf{Generic statements} (e.g., "Vision 2030") are characteristic of machine-generated content.
    \item \textbf{Markdown formatting} present in the text, despite instructions not to use it, suggests machine-generated content.
    \item \textbf{Obvious incorrect information} that seems unlikely for a machine to produce (e.g., "equating MERS COVID") suggests human authorship.
    \item \textbf{Hashtags correctly embedded within the text} point to human-generated content, while those placed at the end of the article suggest machine-generated content.
\end{itemize}

\subsection{Chinese}

\paragraph{Zhihu QA}
We perform both fill-the-gap survey by analyzing whether the gap is filled: Yes or No or Partially, and the manual detection over the improved Zhihu QA responses under the same setting as \tabref{tab:detection-acc}.
For the gap survey across three datasets, issues of 16\% cases are addressed, 35\% are partially addressed, and half is totally not mitigated, as shown in \tabref{tab:fill-gap-survey}.
Detection accuracy for three annotators declines from 99.7\% to 94.3\% on average, respectively from 100\% to 93\%, 99\% to 81\% and 100\% to 99\%.
Annotators carefully read the contents, and then learned the new detection patterns after reading about 25 pairs, and then annotators detect more quickly depending on the following patterns. Overall, after using the improved prompts, model responses become closer to human answers, more challenging than before to discern.

\paragraph{Student Essay}
Three annotators detect using fill the gap survey. Two annotators have similar observations that gaps in 60\% cases are either fully bridged or partially, 40\% remained. Another annotator is more strict, and think problems are not solved on 65\% cases.  

\paragraph{Government Report} 
The new prompts help mitigate the rigid format of model outputs to some extent and enhance output diversity. We observed that in many original examples, the inputs consist of multiple short paragraphs with highly similar lengths. 

However, under the new prompts, the model-generated paragraphs exhibited varying lengths, aligning more closely with typical human writing styles. Despite these improvements, certain repetitive phrasing issues persist and remain unresolved. In our analysis, prompt adjustments proved partially effective. Among 200 test cases, 59 showed a noticeable improvement in output quality, 38 exhibited partial improvement, while no significant changes were observed in 103 cases.

\subsection{English Peer Meta Review}
The new English generations retain many features exhibited by previous generations. Even after modifying the prompts, the LLM remains highly formulaic in its outputs. This may be due to the inherently formulaic nature of the peer review domain.

The first half of the second round of generations typically begin with the phrase ``Based on the reviews ...'' and frequently include bullet points or lists. Although the lists were often formatted with numbers or dashes, instead of traditional bullet points, they tended to contain numerous items, leading to responses that felt rigid and uniform. Repetitive use of the phrase ``Given this I recommend to'' further emphasized the predictable nature of these outputs, making them easily identifiable as machine-generated.

In contrast, the second half of the second round of generations moved away from lists, favoring more narrative, paragraph-based structures. The texts were divided into roughly equal paragraphs, which followed a consistent order: a summary of the paper's content, praise for the paper’s novelty, discussion of reviewer concerns, and a concluding acceptance statement. Variations of phrases such as ``The paper introduces ...'' or ``The manuscript under review ...'' were used to begin the majority of the generations.

Despite these adjustments, the gap between machine-generated and human-written text remains significant. While these improved prompts present a more fluid structure, the length of the texts remains relatively uniform, and the overall structure still lacks the variance and natural inconsistency typically seen in human peer reviews. Human-written reviews tend to exhibit more variability in both text length and structure, resulting in a more organic feel.


\subsection{Hindi News}
In order to address the disparities observed in machine-generated news, an improved prompt was introduced to improve the quality of machine-generated news. The new prompt explicitly provided the original news content as reference, ensuring the inclusion of more factual details, figures, and names. The prompt instructed the model: 
\begin{quote}
\textit{"Here is a news headline: '\{headline\}' and the content: '\{content\}'. Write a machine-generated version of the news based on this headline. Return the news in Hindi and just return the news content in plain formatted text. Don't return anything extra, just return the plain text in Hindi."}
\end{quote}


This approach aimed to reduce previously noted gaps, such as limited content and lack of proper nouns, dates, and contextual richness, while keeping the focus on generating plain, Hindi-formatted news content. Providing original content would also provide an idea of the style of news writing and improve overall coherence, realism, and contextual depth in generated outputs.

Although the improved prompt succeeded in reducing some of the earlier disparities, some of the disparities still existed in the machine-generated text. These included a noticeable underuse of quoted statements and fewer references to names and illustrative examples, both of which are common in human-written news articles. Furthermore, the inclusion of original content for reference led to a significant decrease in classification accuracy, which decreased to 66\% from 85.2\%. The analysis of machine-generated text using improved prompt shows how challenging it can be to improve the style of machine-generated text while keeping it different from human-written text. The results highlight how important it is to carefully design prompts to help models create more natural and human-like text in Hindi.

\subsection{Italian DICE News}
To mitigate easily identifiable cues, we instruct the model to refrain from using any formatting, incorporate details and specific names regarding the event's location, and freely include witness testimonies. Therefore, we incorporate the following additional instructions into the initial generic prompt: 
\begin{quote}
When writing avoid any kind of formatting, do not repeat the title and keep the text informative and not vague, add quotes from witnesses or the police. You don't have to add the date of the event but you can, use at most 300 words. Do not use mark-down formatting.
\end{quote}
or in Italian when testing Anita which is tuned for this language:
\begin{quote}
Quando scrivi evita qualsiasi tipo di formattazione, non ripetere il titolo e mantieni il testo informativo e non vago, aggiungi citazioni da testimoni o dalla polizia. Non devi aggiungere la data dell'evento ma puoi farlo, usa al massimo 300 parole. Non usare formattazione mark-down.
\end{quote}

Refining the prompt makes machine generated text harder to recognize. When annotating the same 300 samples generated with the new prompt, accuracy decreases. Specifically, Anita goes from 88\% to 81\%, Llama-405b from 99\% to 84\% and GPT-4o from 100\% to 85\%. This means that some gaps are bridged by the new prompts, making the machine-generated texts harder to identify. 
With the new prompt, we remove formatting related patterns and we identify others related to writing style:
\begin{itemize}
    \item The generated text occasionally contains words that depart from journalistic writing (noticed at the beginning of the annotation).
    \item Most texts start with one of few prototypical sentences, e.g., \textit{The police ...} and \textit{On that night ...} (noticeable after annotating 10 to 20 samples).
    \item New generations still rarely use quotes, numbers and other specific details (noticeable after annotating 10 to 20 samples).
    \item The model rarely writes in passive voice, which is more usual in Italian than in English (noticeable after annotating 100 samples).
    \item Each model uses consistent writing style that is easily identifiable (noticeable after annotating more than 200 samples).
\end{itemize}

Overall, new generations are partially improved, but there are still some patterns that enable recognition of machine-generated text. The challenges reflect that annotators now need to see more examples than before to summarize these patterns and then leverage them to identify MGT, making zero-shot recognition of MGT harder than before.

\subsection{Japanese News}
Based on the human detection evaluation, we found several distinctive stylistic characteristics in LLM-generated Japanese news with the original prompt, as detailed in \ref{japanese_news_distinctive_clues}.
To prevent LLMs from generating such easily detectable clues, we provide the following improved prompt:

\begin{quote}
    \textit{"次のニュースタイトルに合わせたニュース記事を生成してください。このとき、生成する記事のスタイルには敬体ではなく常体を使用し、またタイトルの言い換えによって記事の冒頭を作成することは避けてください。$\backslash$n ニュースタイトル: \{title\}$\backslash$n ニュース記事: "}
\end{quote}
which indicates
\begin{quote}
    \textit{Please generate a news article that matches the following news title. When creating the article, please use plain form instead of polite form and also avoid generating the beginning of the article by paraphrasing the title.$\backslash$n news title: \{title\}$\backslash$n news article:} 
\end{quote}

By default, while LLMs tend to generate Japanese news articles in a polite form (desu-masu style), human-written news articles are basically in a more plain form (da/dearu style). Therefore, we added instructions to generate articles in a plain form, which is closer to the human writing style in the news domain. Additionally, since LLMs are likely to generate a title at the beginning of the news article, we also included instructions to suppress the task-specific behavior.

\subsection{Kazakh Wikipedia}
Out of 200 annotated examples, our analysis shows that in most cases, the prompt adjustments were effective. We observed 105 instances of ``Yes'', indicating clear improvement, 89 cases of ``Partially'', where some limitations remained, and 6 instances of ``No'', where no noticeable change was observed.

The revised prompt helps reduce repetitive sentence patterns to some extent. The model now produces a more diverse range of sentence structures, partially addressing the issue. However, some predictable phrasing persists, meaning that while variety has improved, formulaic expressions have not been entirely eliminated. The adjustments also successfully increase the inclusion of concrete information, such as dates, names, and specific data points. 



As a result, the generated text now contains more factual details, addressing this issue almost entirely and improving overall informativeness, specificity, and alignment with expected real-world knowledge. Additionally, the prompt helps mitigate excessive flattering language, making the text feel more balanced, neutral, and closer to human-written content. However, some instances of exaggeration still occur, indicating that this issue has only been partially resolved.

The inclusion of Kazakh-specific culturally nuanced details remains the most challenging issue to address. The model defaults to Kazakh-specific references only when the topic is explicitly related to Kazakhstan. When it does not recognize the subject, it often produces factual inaccuracies. For example, when generating text about \textit{Stephen King's It}, the model failed to recognize the name in Kazakh and incorrectly attributed the novel to a prominent Kazakh author, even inventing plot details. This limitation reflects insufficient exposure to Kazakh-specific content, making it difficult to fully resolve through prompt engineering alone.



\subsection{Russian}

\paragraph{Prompt Modifications}
The prompts were adjusted to encourage greater detail and specificity in the generated content:
\begin{itemize}
\item \textbf{Academic summaries:} The simple prompt asked the model to generate a summary without a title based on a given topic and title. The improved prompt added instructions to include details about results, experiments, numbers, and references to other works.
\item \textbf{News articles:} The simple prompt instructed the model to generate a news article based on the given title and topic from the Lenta.ru website without a heading. The improved prompt emphasized including as many details as possible, such as names, numbers, and dates, and encouraged citing original sources like RIA Novosti, CNN, or BBC.
\end{itemize}
After implementing the improved prompts, we observed a decrease in the accuracy of machine-generated text detection. Specifically, for news articles, accuracy dropped from 100\% to 86.5\%, while for summaries, it decreased from 80\% to 69\%. This suggests that the refined prompt instructions successfully made machine-generated text harder to distinguish from human-written text.

\paragraph{Observations and Patterns}

The enhanced prompts led to increased textual complexity and information density, which made it more difficult to detect machine-generated text. However, new artefacts appear, which became noticeable after reviewing 50–70 samples, particularly in recurring phrasing, tone consistency, and subtle structural regularities:

\begin{itemize}
\item \textbf{Repetitive references:} The generated summaries often cited the same set of names (e.g., \textit{Smith} and \textit{Jones}), while news articles frequently referred to a limited set of news outlets (e.g., \textit{RIA Novosti}).
\item \textbf{Text length:} Human-written news articles were generally shorter than machine-generated ones, which tended to be overly detailed.
\item \textbf{Standardized introduction phrases:} Summaries frequently begin with a predictable opening sentence, such as \textit{``The article discusses ...''}.
\item \textbf{Identifiable uniqueness:} If a summary contained references to names outside the repetitive set or a news article cited a unique news outlet, it was more likely to be human-written. Otherwise, detection became significantly harder.
\end{itemize}


While the refined prompts made detection more difficult, annotators who analyzed a sufficient number of samples could still identify machine-generated text based on these emerging patterns and subtle stylistic consistencies that persist across different outputs and domains.

\begin{figure*}[t!]
    \centering
    \includegraphics[scale=0.6]{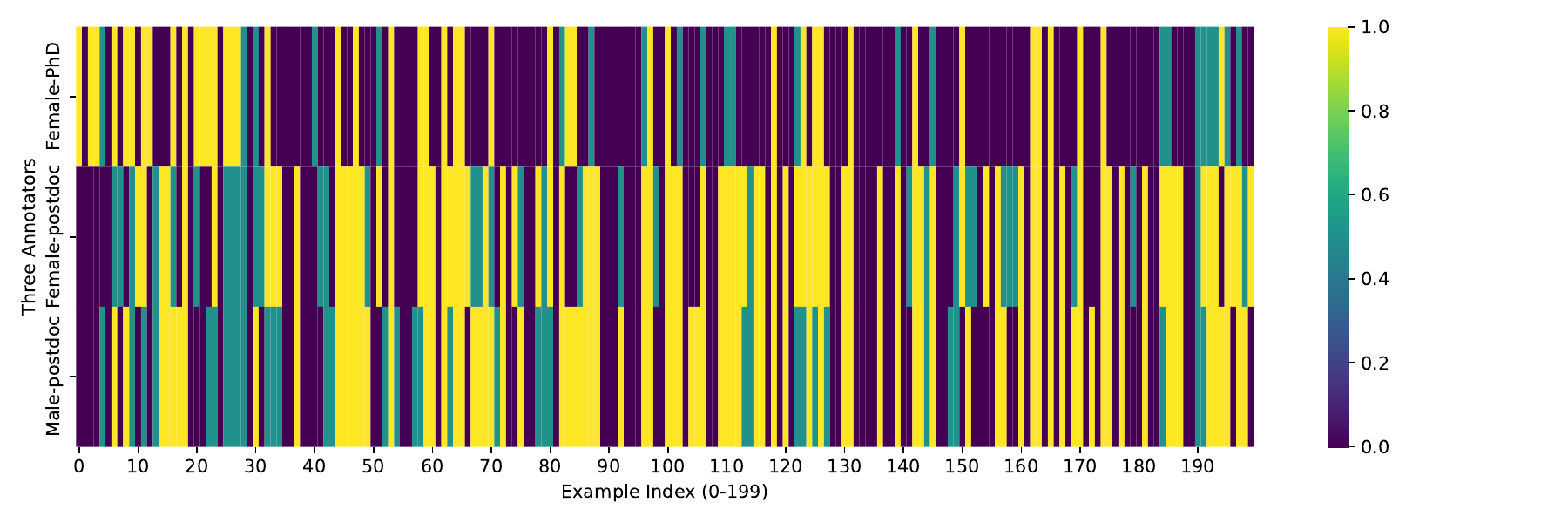}
    \caption{Three annotator agreement on Chinese essays regarding whether the improved prompts mitigate the gap between human text and machine-generated text.}
    \label{fig:iaa-heatmap}
\end{figure*}

\subsection{Vietnamese}
\selectlanguage{Vietnamese}

To address the disparities in these two domains, we introduced a new prompt for each domain to enhance machine-generated text and better align outputs with human writing patterns.

For Vietnamese news, we use the prompt:
\begin{quote}
    \textit{Bạn là một nhà báo Việt Nam chuyên viết những phần đầu đề cho các bài báo bằng cách sử dụng tiêu đề của chúng. Lưu ý rằng, hãy đưa ra các thông tin cụ thể và chính xác như mốc thời gian, địa điểm, các tình tiết, chi tiết, đồng thời viết văn phong dưới dạng tóm lược cho phần mở đầu của bài báo. Hãy viết cho tôi một đoạn đầu đề cho bài báo có tiêu đề dưới đây:}
\end{quote}
which means
\begin{quote}
    \textit{"You are a Vietnamese journalist who writes headlines for articles using their titles. Please be specific and precise in giving information such as time, place, circumstance, and details, and write a summary for the introduction of the article. Please write me a headline for the article with the following title:"} 
\end{quote}
This prompt is used to improve writing quality by providing more detailed information, such as dates and times, rather than merely offering general information in the news title. While this information may sometimes be incorrect, its specificity can persuade readers by appearing more credible (e.g., \textit{"In May 2020, Mr. A, the president of company X, stated that..."}).

For Wikipedia articles, we refined the prompt to ensure that machine-generated text includes only information that is appropriate for Wikipedia. We also asked to write only the introductory information, without delving into further details. Our new prompt for Vietnamese Wikipedia is
\begin{quote}
    \textit{"Bạn là một nhà đóng góp cho Wikipedia tiếng Việt. Hãy viết cho tôi một đoạn giới thiệu ngắn gọn bằng tiếng Việt về chủ thể bên dưới để đăng trên trang Wikipedia. Hãy cố gắng hiểu về chủ đề, và viết ra một đoạn giới thiệu chứa các thông tin mà người dùng thường tìm kiếm trên Wikipedia. Hãy cố gắng viết ra các đoạn văn bản giống người viết nhất có thể. Chỉ đưa ra đoạn giới thiệu, không đưa ra thêm các thông tin khác. Lưu ý rằng bạn phải giữ độ dài văn bản trong khoảng \texttt{word\_count} từ, không được viết dài hơn."}
\end{quote}
which means:
\begin{quote}
    \textit{"You are a contributor to Vietnamese Wikipedia. Please write me a short introduction in Vietnamese about the subject below to post on the Wikipedia page. Try to understand the topic, and write an introduction that contains information that users often search for on Wikipedia. Try to write the text as humanly as possible. Provide only the introduction, do not provide any other information. Note that you must keep the text length within \texttt{word\_count} words, do not write longer."}
\end{quote}


Generally, these improved prompts aim to imbue AI-generated text with characteristics similar to human-written passages, enabling AI to produce more human-like text. Indeed, humans find it more difficult to distinguish between the two passages, as they are more confused while reading them and struggle to determine which one is machine-generated. The contextual gap between human and AI seems to have narrowed significantly.


However, as noted previously, for news articles, the machine may hallucinate and generate incorrect timestamps for events. Additionally, the writing style of the machine is somewhat unique and consistent across records, whereas human-written text tends to exhibit greater variation in structure and length. This distinction remains a key factor for humans when identifying whether a passage is machine-generated or written by a human.

\selectlanguage{English}

\begin{table*}[t!]
    \centering
    \resizebox{\textwidth}{!}{
    \begin{tabular}{ll  p{9cm}  p{9cm}}
    \toprule
    \textbf{Language} & \textbf{Source} & \textbf{Original Prompt} & \textbf{Improved Prompt} \\
    \midrule
    Arabic & Dialect Tweet & [translated from AR] Write a tweet in '\{dialect\}'. [English Prompt] Write a random tweet in '\{dialect\}' & [translated from AR] Write a random tweet in Arabic. Use dialect '\{dialect\}', express emotions and human experience. [English Prompt] Generate a random tweet in Arabic. Use dialect '\{dialect\}', use human emotions and experience. Output the tweet only. \\
    Arabic & EASC Summary& [translated from AR] Summarize this article while preserving the key points, ensuring conciseness and accuracy: '\{article\}'.& 
    [translated from AR] Summarize the following article while attempting to simulate a human with intellectual or religious beliefs, ensuring accuracy and conciseness, as per the average human level: '\{article\}'.\\
    Arabic & Youm7 News & [translated from AR] Paraphrase the following news article in clear, clear language, keeping the key information and ideas accurate. Make sure to arrange paragraphs logically and present ideas in an understandable sequence, while using fluent, easy-to-read Arabic. & [translated from AR] Rewrite the provided news article with a refined, well-structured, and sophisticated style that enhances clarity, depth, and coherence. Ensure the article maintains journalistic integrity while improving logical flow, sentence structure, and readability. The revised version should elevate the narrative by incorporating precise language, nuanced transitions, and a compelling tone appropriate for a professional audience. Maintain factual accuracy, emphasize key points effectively, and optimize the article’s structure to enhance engagement and comprehension. Additionally, refine the coherence between paragraphs, eliminate redundancy, and ensure seamless progression of ideas.\\
    Arabic & SANAD News & [translated from AR] You are a professional news writer. Write an article in Arabic consisting of approximately \{word\_count\_rounded\} words. The article is titled: '\{title\}', assume that the source is: '\{source\}', and the general topic falls under '\{topic\_arabic\}'. & 
    [translated from AR] Act as if you are a professional news writer and write an article in Arabic consisting of approximately \{word\_count\_rounded\} words, titled \{title\}. Ensure that the article is rich in information and precise details, including numbers, dates, and quantitative data when relevant. Make sure to include links, phone numbers, or currency exchange rates when necessary. Use some specialized English terms if the context requires it, and avoid relying solely on generic phrases like "This development comes amid a global trend..." without supporting events or details. Maintain a clear chronological order that reflects the logical sequence of events. Ensure that the text is written as a single paragraph regardless of its length, without dividing it into multiple paragraphs. Keep the formatting imperfect (e.g., uneven spacing) to reflect the nature of human writing, and feel free to use hashtags when needed. Please note that the topic generally falls under the category of \{topic\_arabic\}. Please do not use Markdown at all.\\

     \midrule
    Hindi & News & 
    [translated from Hindi] Here is a news headline: '{headline}' and the content: '{content}'. Write a machine-generated version of the news based on this headline. & 
    [translated from Hindi] Here is a news headline: '{headline}' and its content: '{content}'. Generate a machine-written version of the news based on this headline. Return the news in Hindi, formatted as plain text. Do not include any additional text—just return the generated news content in Hindi. \\
    \midrule
    Italian & DICE News
    & \textbf{System:} You are an Italian journalist writing for a national newspaper focusing on criminal events happening in the area surrounding Modena. \newline \textbf{User:} Write a piece of news in Italian, that will appear in a local Italian newspaper and that has the following title: & \textbf{System:} You are an Italian journalist writing for a national newspaper focusing on criminal events happening in the area surrounding Modena. \newline \textbf{User:} In writing avoid any kind of formatting, do not repeat the title and keep the text informative and not vague. You don't have to add the date of the event but you can \\
    \midrule
    Japanese & News & Please generate a Japanese news article that matches the following news title.$\backslash$n news title: \{title\}$\backslash$n news article: & Please generate a Japanese news article that matches the following news title. When creating the article, please use plain form instead of polite form, and avoid generating the beginning of the article by paraphrasing the title.$\backslash$n news title: \{title\}$\backslash$n news article: \\
     \midrule
    Kazakh & Wikipedia & Please, write one paragraph about the following topic in Kazakh:  \texttt{\{topic\}}. \newline \foreignlanguage{russian}{Мына тақырып туралы бір абзац жазыңыз: \texttt{{тақырып}}.} & Please, write one paragraph in Kazakh about the following topic: \texttt{\{topic\}}. Avoid repetitive sentence structures and predictable phrasing. When applicable, include concrete facts such as specific numbers, dates, or historical references. Avoid overly flattering or exaggerated language, and instead focus on delivering clear, informative, and relevant content. \newline  \foreignlanguage{russian}{Мына тақырып туралы бір абзац жазыңыз: \texttt{{тақырып}}. Қайталағыш сөйлем құрылымдарынан және болжамды сөздерден аулақ болыңыз. Қажет болған жағдайда нақты деректерді, мысалы, нақты сандарды, даталарды немесе тарихи сілтемелерді қосыңыз. Артық мақтау немесе асыра сілтеуден аулақ болыңыз, орнына анық, ақпаратты және сәйкес мазмұнды жеткізуге назар аударыңыз.} \\
    
    \bottomrule
    \end{tabular}
   }
\end{table*}

\begin{table*}[t!]
    \centering
    \resizebox{\textwidth}{!}{
    \begin{tabular}{ll  p{9cm}  p{9cm}}
    \toprule
    \textbf{Language} & \textbf{Source} & \textbf{Original Prompt} & \textbf{Improved Prompt} \\
    \midrule
    English & Peersum & 
   Generate a meta review based on the reviews' opinions and authors' rebuttal to make the final decision on whether the paper should be accepted: \texttt{\{Reviews\}} $\backslash $n$\backslash$n Meta review:  & 
    Generate a meta review based on the reviews' opinions and authors' rebuttal to make the final decision on whether the paper should be accepted: \texttt{\{Reviews\}}$\backslash $n$\backslash$n. When generating, don't use rigid format or structure such as ``Decision: XX" or headings such as ``Strengths", ``weaknesses" or bullet points. $\backslash $n$\backslash$n Meta review: \\
    \midrule
    Chinese & Zhihu-QA & [translated from Chinese] Imagine you are a rational and analytical Zhihu user. Your objective is to answer question in a clear, objective, and logical manner. Please provide a thorough and well-supported answer. \newline \cn{假设你是一位理性分析的知乎用户，你的目标是以客观、逻辑的方式回答以下问题。请以理性分析者的身份，给出详细、有据可查的回答。 } & [translated from Chinese] Imagine you are a rational and analytical Zhihu expert. Your goal is to provide objective, logical, and detailed answers while strictly adhering to Zhihu’s style. Keep responses concise, avoiding excessive politeness or formality. Based on the question's professionalism, start with 'Thank you for the invitation' or a similar phrase. If relevant, @ professionals to enhance credibility. \newline \cn{假设你是一位理性分析的知乎专家，你精通提问者所提出的问题。 你的目标是以客观、逻辑的方式回答问题。请给出详细、有据可查的回答。回答请严格符合知乎风格，避免长答案和过于礼貌正式的答案。根据问题的专业性，在开头回复“谢邀”或同义词，在必要时请在回答中@一些行业专业人士。}  \\
    Chinese & Student essay & [translated from Chinese] As a Gaokao student, you must write an 800-word essay that is logical, well-structured, and clearly argued, with sufficient evidence. Express unique insights on social, life, or philosophical issues in formal, elegant language, avoiding colloquialisms. Ensure smooth transitions and a complete structure, with a conclusion that reinforces and elevates the theme. \cn{你是一名正在参加高考的学生，现在需要完成一篇800字左右的作文。请根据题目要求，深刻思考，并清晰表达你的观点。文章应有逻辑性和层次感，论点明确、论据充分，能够展示出你对社会、生活或哲理问题的独到见解。注意语言要规范、优美，避免口语化表达，确保文章结构完整，段落之间过渡自然。结尾部分应对文章的主题进行总结和升华，展示你对问题的深入思考。} & [translated from Chinese] As a Gaokao student, you are required to write an 800-word essay expressing your views clearly and logically, with well-structured arguments and supporting evidence. Reflect on social, life, or philosophical topics, using formal and elegant language while avoiding colloquial expressions. Ensure smooth transitions between paragraphs and a conclusion that deepens the theme. Avoid rigid structures like “firstly” and “secondly,” and if writing narrative, use sincere and emotionally resonant language. Do not include a title or heading, and write in Chinese only. \newline \cn{你是一名正在参加高考的学生，现在需要完成一篇800字左右的作文。请根据题目要求，深刻思考，并清晰表达你的观点。文章应有逻辑性和层次感，论点明确、论据充分，能够展示出你对社会、生活或哲理问题的独到见解。注意语言规范、优美，避免口语化表达，确保文章结构完整，段落之间过渡自然。结尾应对文章的主题进行总结和升华，展示你对问题的深入思考。使用多样化的表达，避免过于结构化的表述，例如"首先、其次、然后"等；如果是记叙文，请使用情感丰富的真实表达。请不要输出任何题目或标题，直接开始写作。请全程使用中文。} \\
    Chinese & Government Report &[translated from Chinese] Please continue writing the full article based on the provided introduction. The original article starts with: \{head\}  \cn{请根据文章的开头续写完整的文章，原文章开头为: \{head\}} & [translated from Chinese] Please continue writing the full article based on the provided introduction: {head}. Please make the article as long as possible without generating symbols like **. Do not generate any content unrelated to the article. The original article starts with: \{head\} \newline \cn{请根据文章的开头续写完整的文章，请让文章尽可能的长,并且不生成**这类符号。请不要生成文章之外的其他内容。原文章开头：\{head\}}  \\

    \midrule
    Russian & News & Write a news article on the topic \texttt{\{topic\}} from the lenta.ru website using the title \texttt{\{title\}}. You must generate news without a heading. News: \newline \foreignlanguage{russian}{Напиши новость в области \texttt{\{topic\}} с сайта lenta.ru используя заголовок \texttt{\{title\}}. Ты должен генерировать новость без заголовка. Новость:}
    & Write a news article on the topic \texttt{\{topic\}} from the lenta.ru website using the title \texttt{\{title\}}. You must generate news without a heading. Add as many details as possible, such as names, numbers, dates, etc. You should also refer to the original source of the information (for example, RIA Novosti, CNN, BBC, etc.). News: \newline \foreignlanguage{russian}{Напиши новость в области \texttt{\{topic\}} с сайта lenta.ru используя заголовок \texttt{\{title\}}. Ты должен генерировать новость без заголовка. Добавь побольше деталей, такие как имена, числа, даты и тд. Так же ты должен сослаться на первоисточник информации (например, РИА Новости, CNN, BBC и т.д.). Новость:}\\
    Russian & Academic summary & Write a summary for an article in the \texttt{\{topic\}} topic using the title \texttt{\{title\}}. You must generate a summary without a title. Contents: \newline \foreignlanguage{russian}{Напиши краткое содержание для статьи в области \texttt{\{topic\}} используя заголовок \texttt{\{title\}}. Ты должен генерировать краткое содержание без заголовка. Содержание:}
    & Write a summary for an article in the \texttt{\{topic\}} topic using the title \texttt{\{title\}}. You should generate a summary without a title. Add details about the results, experiments, numbers, references to other work, etc. Contents: \newline \foreignlanguage{russian}{Напиши краткое содержание для статьи в области \texttt{\{topic\}} используя заголовок \texttt{\{topic\}}. Ты должен генерировать краткое содержание без заголовка. Добавь деталей про результаты, эксперименты, числа, ссылки на другие работы и т.д. Содержание:}\\

    \bottomrule
    \end{tabular}
   }
\end{table*}

\begin{table*}[t!]
    \centering
    \resizebox{\textwidth}{!}{
    \begin{tabular}{ll  p{9cm}  p{9cm}}
    \toprule
    \textbf{Language} & \textbf{Source} & \textbf{Original Prompt} & \textbf{Improved Prompt} \\
    \midrule
    Vietnamese & Wikipedia & You are a contributor to Vietnamese Wikipedia. Please write me a brief introduction in Vietnamese about the subject below to post on the Wikipedia page. Note that it must be written within \texttt{word\_count} words: & You are a contributor to Vietnamese Wikipedia. Please write me a short introduction in Vietnamese about the subject below to post on the Wikipedia page. Try to understand the topic, and write an introduction that contains information that users often search for on Wikipedia. Try to write the text as humanly as possible. Provide only the introduction, do not provide any other information. Note that you must keep the text length within \texttt{word\_count} words, do not write longer.\\
    Vietnamese & News & You are a Vietnamese journalist who writes summaries for articles using their headlines. Please write me a summary of the article with the following headline: & You are a Vietnamese journalist who writes headlines for articles using their titles. Please be specific and precise in giving information such as time, place, circumstance, and details, and write a summary for the introduction of the article. Please write me a headline for the article with the following title:\\
    \bottomrule
    \end{tabular}
   }
    \caption{Original prompts vs. improved prompts used to fill the gap between human and machine text, imitating human style and writing convention.}
    \label{tab:ori-improved-prompts}
\end{table*}

\subsection{Subjectivity in Fill-the-gap Survey}
\label{app:subjectivity-test}

To examine the subjectivity of this survey, i.e., variation between annotators when evaluating the same examples, we asked three annotators who previously detected Chinese student essays to assess the same set of newly generated essays and measure inter-annotator agreement over 200 cases.
\figref{fig:iaa-heatmap} shows that the two postdoc annotators exhibit higher correlation, whereas the female PhD annotator demonstrates significant disagreement.
This suggests annotators' backgrounds influence results, and individual biases are non-negligible factors.
To obtain more objective results, we conducted a second detection round on the new generations, as detection accuracy provides a more objective measure.

\section{Automatic Detection Data and Results}
\label{app:automaticdetectiondata}
\tabref{tab:multilingual_prompt_dist} presents the distribution of data used in automatic MGT detection, and the results are shown in \figref{fig:auto-acc-diff}. 19 among 26 automatic detection approaches demonstrate lower detection accuracy on the improved generations produced by the adjusted prompts.


\begin{table*}[ht]
\centering
\scalebox{0.7}{
\begin{tabular}{>{\raggedright\arraybackslash}p{3cm}|>{\centering\arraybackslash}p{2cm}>{\raggedleft\arraybackslash}p{1.9cm}>{\raggedleft\arraybackslash}p{1.6cm}>{\raggedright\arraybackslash}p{12cm}}
\hline
\rule{0pt}{2.5ex}Source / Domain & Language & \# Improved & \# Original & LLM Generator List \rule[-1.2ex]{0pt}{0pt}\\
\hline
\rule{0pt}{2.5ex}QA & Chinese & 3422 & 9842 & GPT-4o (3421), GPT-4o-mini (6845), GPT-4o-2024-05-13 (2998)\rule[-1.2ex]{0pt}{0pt}\\
\hline
\rule{0pt}{2.5ex}Essay & Chinese & 849 & 702 & Claude-3-5-Sonnet (773), GLM-4-9B-Chat (778)\rule[-1.2ex]{0pt}{0pt}\\
\hline
\rule{0pt}{2.5ex}GovReport & Chinese & 16776 & 0 & Baichuan2-13B-Chat (5521), ChatGLM3-6B (5359), GPT-4o (5896) \rule[-1.2ex]{0pt}{0pt}\\
\hline
\rule{0pt}{2.5ex}News & Hindi & 0 & 1199 & GPT-4 (600), Human (599)\rule[-1.2ex]{0pt}{0pt}\\
\hline
\rule{0pt}{2.5ex}News & Japanese & 0 & 300 & GPT-4o (300) \rule[-1.2ex]{0pt}{0pt}\\
\hline
\rule{0pt}{2.5ex}News & Russian & 5915 & 600 & GPT-4o (3921), Vikhrmodels/Vikhr-Nemo-12B-Instruct-R-21-09-24 (3224)\rule[-1.2ex]{0pt}{0pt}\\
\hline
\rule{0pt}{2.5ex}Sunmary & Arabic & 153 & 153 & GPT-4o (306)\rule[-1.2ex]{0pt}{0pt}\\
\hline
\rule{0pt}{2.5ex}Sunmary & Russian & 5944 & 585 & Vikhrmodels/Vikhr-Nemo-12B-Instruct-R-21-09-24 (3279), GPT-4o (3300)\rule[-1.2ex]{0pt}{0pt}\\
\hline
\rule{0pt}{2.5ex}Tweets & Arabic & 2800 & 214 & GPT-4-Turbo (1400), GPT-4 (1545), Qwen2.5 72B (69)\rule[-1.2ex]{0pt}{0pt}\\
\hline
\rule{0pt}{2.5ex}Total & -- & 32,487 & 17,017 & -- \rule[-1.2ex]{0pt}{0pt}\\
\hline
\end{tabular}
}
\caption{Statistics of data used in automatic detection: generations using original and the improved prompts.}
\label{tab:multilingual_prompt_dist}
\end{table*}

\section{Human Preference}
\subsection{Why do Human Preference Strongly Differ in Zhihu QA?}
\label{app:humanpreference}
We summarized three major reasons.

\textbf{Mean-spirited human Zhihu answers.}
One of the female annotators prefers responses that are sincere, authentic, and uplifting. However, many human-written responses tend to show the dark aspects of society by often adopting a strongly negative tone through harsh, sarcastic, or offensive remarks targeting specific groups. While these responses may contain some facts, they present a narrow, subjective perspective rather than a comprehensive view. In such cases, a more positive or neutral response is generally perceived as more mature and reasonable, aligning with Chinese cultural norms of expressing individual opinions in a mild and modest manner.  

This does not imply that machine-generated responses are ideal. When asked for personal opinions, LLMs typically provide generic, average viewpoints that lack depth, wisdom, and inspiration. Human annotators prefer personalized, insightful reflections and suggestions informed by real experiences and critical thinking, rather than generic statements that reiterate common knowledge. 

This may stem from lack of personal experiences and inability to internalize knowledge as a coherent philosophical perspective or behavioral framework.

Moreover, human answers would extend the answer from the current topic to other relevant topics, rather than only responding to this question.

They can tell from the history and naturally incorporate personal thoughts into it, while models do not have this ability or nature.

\textbf{Humans more trust human suggestions.}
For questions asking for clinical suggestions, graduate program application experience, recommendation of restaurants, educational institution, tutoring courses, and teacher in a city, town or even a district, humans tend to trust more on human answers than model generations.
These questions require either expert knowledge or real personal experience.
Compared with model generic opinion and suggestions, humans offer advice in a way of real professional expertise or share firsthand experience with more details, feeling and suggestions, preferred by individuals.
For example, \textit{How many students did Tongda Students and Kexing Education Tutoring Institution get admitted to top universities this year?} Kexing Education and Tongda Students are two tutoring schools helping high school students with admissions. This internal information is only known to a small group, making LLM outputs less trustworthy and often factually incorrect.


\textbf{Emotional and complex relationship issues are difficult for models.}

Sometimes it is difficult for models to understand complex relationships and emotional issues and provide practical suggestions.
For example, \textit{I have 87 days left until the college entrance exam as a student who attended this exam the second time, and my girlfriend, who is in university, and we have been together for over 450 days. When I went to see her, she told me she was exhausted and that I couldn't provide the support she needed. She said that I was not someone who could stand by her through tough times. Since this was my first relationship, I don't know how to do. She broke up with me decisively.}
In such a scenario, human responses are often more realistic and practical, drawing from personal experience, expressing empathy, and offering actionable advice, though phrasing may sometimes be less polished or emotionally nuanced. They also tend to acknowledge emotional complexity, validate feelings, and provide context-aware guidance grounded in lived experiences and interpersonal understanding.

\subsection{Preference Distributions Vary Across Individuals}
The annotators for Russian and Arabic exhibited similar preference distributions, whereas large differences occurred for Chinese QA.
For example, in the Chinese Zhihu QA, the second annotator selected only seven human-written texts, while the fourth one chose 284 (95\%). In the QA-emo, the first and second annotators preferred human-written responses, whereas the third and the fourth favored machine-generated text.
This variance shows the charm of collecting personal preferences and then optimizing models to align with individual philosophies. 
Our preference annotations can serve as a valuable resource for investigating the relationship between annotator characteristics (e.g.,\ MBTI personality, gender, and age) and preferences. Also, the data can guide models to match individual preferences in multilingual contexts.

\end{document}